\pdfoutput=1
\documentclass[11pt]{article}
\usepackage[preprint]{acl}

\usepackage{times}
\usepackage{latexsym}
\usepackage{hyperref}
\usepackage{amssymb}
\usepackage{pifont}
\usepackage{color}
\definecolor{ao(english)}{rgb}{0.0, 0.5, 0.0}
\usepackage{colortbl, xcolor}
\usepackage{xparse}
\usepackage{arydshln}
\usepackage{booktabs}
\usepackage{array}
\usepackage{multirow}
\usepackage{graphicx}
\usepackage{makecell}
\usepackage{amsmath}
\usepackage{tabularx}
\usepackage{xcolor}
\usepackage{tcolorbox}
\usepackage{tikz}
\usepackage{enumitem}
\usepackage{amssymb}
\usepackage{subcaption}
\usepackage{adjustbox}
\newcommand{\notdingcheck}{
    \begin{tikzpicture}[scale=0.3, baseline=-0.2em]
        \node at (0,0) {\color{blue}\ding{51}}; 
        \draw[line width=0.4mm, color=blue!60!black] (-0.2,0.3) -- (0.6,-0.3);
    \end{tikzpicture}%
}

\NewExpandableDocumentCommand\XGap{m}{\noalign{\vskip #1}}
\NewExpandableDocumentCommand\Gap{}{\XGap{3pt}}

\usepackage[T1]{fontenc}

\usepackage[utf8]{inputenc}
\usepackage{microtype}

\usepackage{inconsolata}
\usepackage{graphicx}

\newcommand{\ours}{FCMR}
\newcommand{\oursfullname}{Financial Cross-Modal Multi-Hop Reasoning}

\newcommand{\ourGen}{CMRGen}
\newcommand{\ourGenfullname}{Cross-Modal Multi-Hop Reasoning Generator}

\title{\ours{}: Robust Evaluation of Financial Cross-Modal Multi-Hop Reasoning}

\author{Seunghee Kim$^{\dagger}$, Changhyeon Kim$^{\dagger}$, Taeuk Kim$^*$ \\
Hanyang University, Seoul, Republic of Korea \\
{\tt \{gyg9325,livex,kimtaeuk\}@hanyang.ac.kr}}

\begin{document}
\maketitle
\begin{abstract}

\renewcommand{\thefootnote}{}
\footnote{$^{\dagger}$Equal contribution. $^*$Corresponding author.
\addtocounter{footnote}{-1}}
Real-world decision-making often requires integrating and reasoning over information from multiple modalities.
While recent multimodal large language models (MLLMs) have shown promise in such tasks, their ability to perform multi-hop reasoning across diverse sources remains insufficiently evaluated.
Existing benchmarks, such as MMQA, face challenges due to (1) data contamination and (2) a lack of complex queries that necessitate operations across more than two modalities, hindering accurate performance assessment.
To address this, we present \oursfullname{} (\ours{}), a benchmark created to analyze the reasoning capabilities of MLLMs by urging them to combine information from textual reports, tables, and charts within the financial domain.
\ours{} is categorized into three difficulty levels---Easy, Medium, and Hard---facilitating a step-by-step evaluation.
In particular, problems at the Hard level require precise cross-modal three-hop reasoning and are designed to prevent the disregard of any modality.
Experiments on this new benchmark reveal that even state-of-the-art MLLMs struggle, with the best-performing model (Claude 3.5 Sonnet) achieving only 30.4\% accuracy on the most challenging tier.
We also conduct analysis to provide insights into the inner workings of the models, including the discovery of a critical bottleneck in the information retrieval phase.
\end{abstract}

\begin{figure}[t]  
\centering
\includegraphics[width=0.9\columnwidth, keepaspectratio]{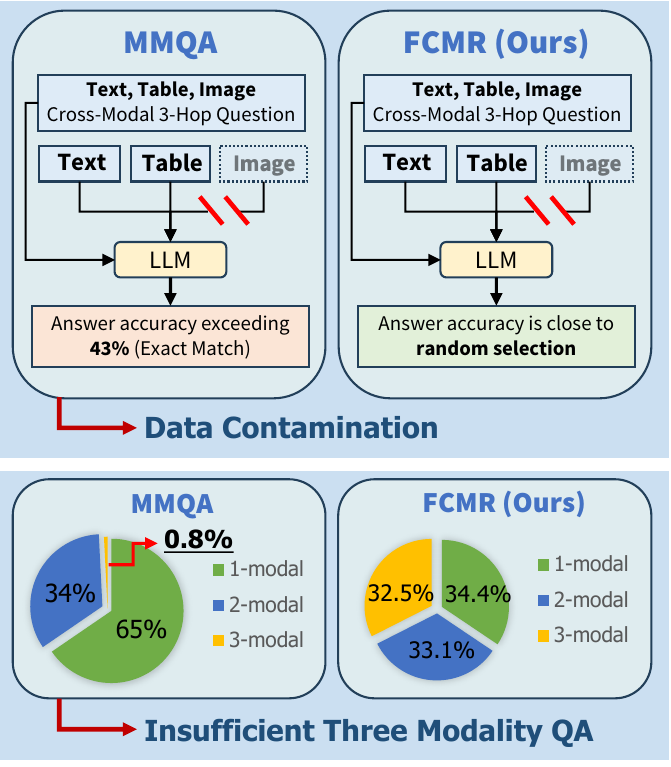}
  \caption{Prior benchmarks for cross-modal multi-hop reasoning, such as MMQA \cite{talmor2021multimodalqa}, exhibit some flaws. MMQA’s cross-modal three-hop questions are often solvable without images, and its complexity is limited, with only 0.8\% of instances having three modalities. In contrast, \ours{} addresses these issues.}
\label{fig:mmqa_vs_ours}
\end{figure}

\begin{figure*}[t]
\centering
\includegraphics[width=0.95\textwidth, keepaspectratio]{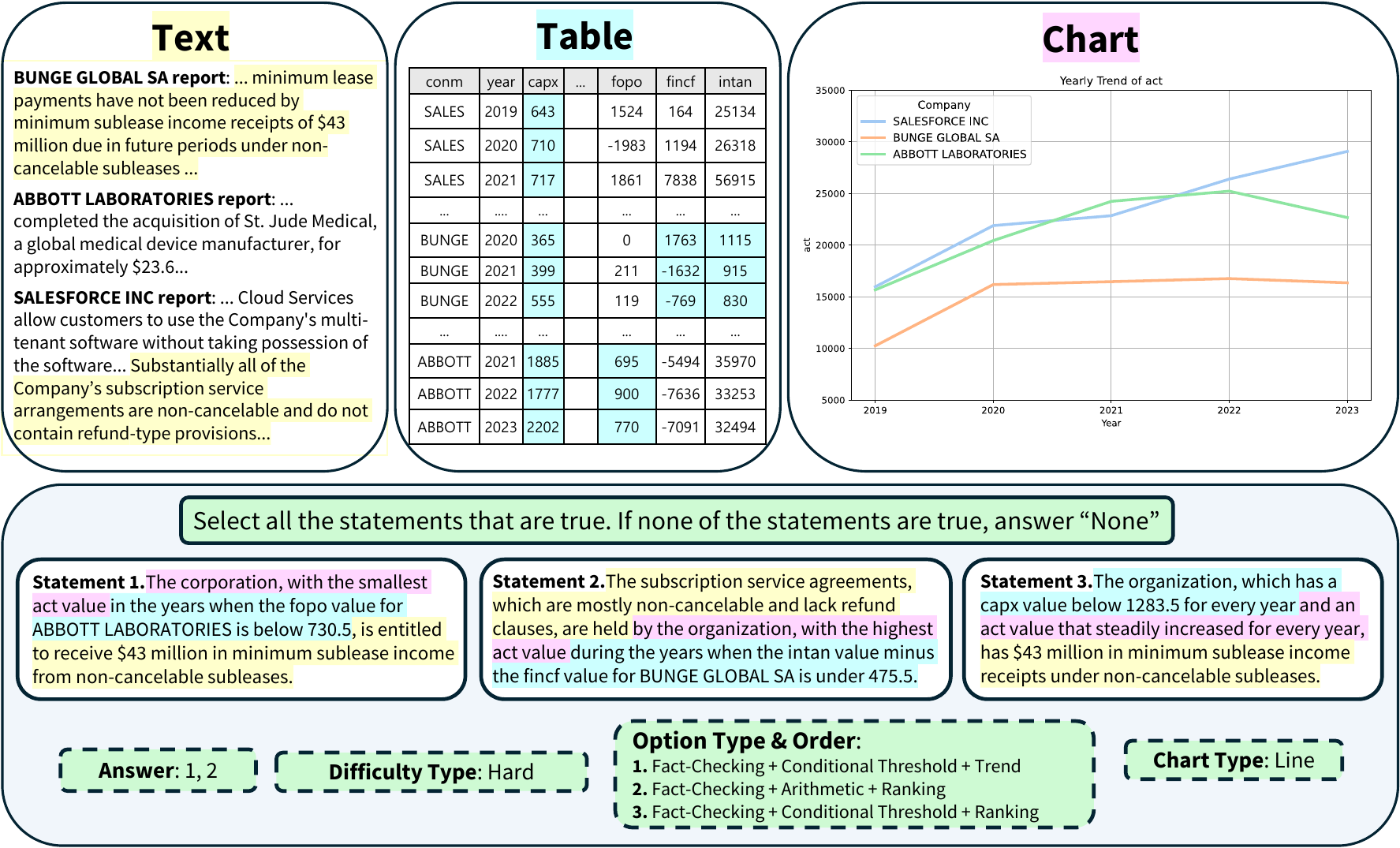}
  \caption{An example from \ours{} at the Hard difficulty level, where all statements require cross-modal three-hop reasoning. Highlights in \textcolor{yellow}{yellow}, \textcolor{cyan}{cyan}, and \textcolor{pink}{pink} denote information from text, tables, and charts, respectively. The model must list all true statements and is correct only if its final prediction (``1, 2'' in this case) is accurate. Information within dashed lines is used only for data generation and excluded from actual instances.}
\label{fig:fcmu_example}
\end{figure*}

\section{Introduction}
Despite the recent progress in AI \cite{touvron2023llamaopenefficientfoundation, anthropic2024claude, openai2024gpt4technicalreport}, developing systems capable of human-level reasoning remains a challenge.
Human cognition involves integrating information from multiple modalities to comprehend and make decisions.
One domain that requires such a comprehensive understanding is finance {\cite{xie2024openfinllmsopenmultimodallarge}, where analysts must simultaneously examine textual reports, tabular data (e.g., balance sheets), and visual data (e.g., charts).
For example, verifying the statement from Figure \ref{fig:fcmu_example}---\textit{``The corporation, with the smallest act value in the years when the fopo value for ABBOTT LABORATORIES is below 730.5, is entitled to receive \$43 million in minimum sublease income from non-cancelable subleases.''}---one must consider all the clues provided by each source, an ability we refer to as 
\textbf{cross-modal multi-hop reasoning}. 

While the literature \cite{chen2020hybridqa, hannan2020manymodalqa,talmor2021multimodalqa,chang2022webqa} presents initial attempts to evaluate the cross-modal multi-hop reasoning capabilities of multimodal large language models (MLLMs), these efforts exhibit several critical shortcomings that undermine their robustness.
First, the heavy reliance on Wikipedia as the foundation for most benchmarks raises concerns.
As Wikipedia is widely known to be a key resource in the pretraining of many recent models, evaluations using Wikipedia-based datasets risk introducing inherent biases. 
These biases may skew results in favor of models that simply recall memorized knowledge, rather than accurately assessing reasoning abilities on unseen data.
Moreover, the scope of validation needs to expand to encompass professional domains, such as finance and science.

Second, current benchmarks are largely focused on testing straightforward problems, such as single- and two-hop reasoning.
As shown in Figure \ref{fig:mmqa_vs_ours}, MMQA \cite{talmor2021multimodalqa}---one of the leading benchmarks in this field---features only about 0.8\% of the queries that explicitly require \textit{three-hop} cross-modal reasoning.
Furthermore, in preliminary experiments, we discovered that GPT-4o \cite{openai2024gpt4technicalreport} can solve the MMQA’s most challenging problems with a 43\% exact match accuracy, even without access to visual clues.
This result highlights the urgent need to establish a higher standard for evaluating cross-modal multi-hop understanding in a more robust and reliable manner.


In this work, we propose \textbf{\oursfullname{} (\ours{})},\footnote{Our dataset is available at \url{https://github.com/HYU-NLP/FCMR}.} a novel benchmark designed to address the limitations of existing datasets in cross-modal multi-hop reasoning.
\ours{} provides multiple-choice QA samples that test the integration of facts from text, tables, and charts.
It consists of three levels of difficulty: Easy, Medium, and Hard.
As in Figure \ref{fig:mmqa_vs_ours}, all instances in \ours{} necessitate understanding three modalities to be answered correctly.
In addition, problems at the Hard level explicitly demand cross-modal three-hop reasoning, making them more challenging (Figure \ref{fig:fcmu_example}).
As \ours{} is built using data sources from the financial domain, it is relatively free from the risk of data contamination.  

Experiments on \ours{} confirm that it poses challenges even for state-of-the-art MLLMs, e.g., GPT-4o and Claude 3.5 Sonnet, encouraging research efforts to develop systems capable of reasoning across multiple modalities.
For analysis, we define four procedures of cross-modal multi-hop reasoning---Planning, Modality Identification, Information Retrieval, and Information Reasoning---and probe diverse models.
We reveal that models particularly struggle with the Information Retrieval phase, implying that MLLMs often fail to extract information from a specific modality, even when they successfully identify where the required facts are located.
We also report findings from additional analyses, including the observation that MLLMs have difficulty counting negative numbers.

\begin{table}[t]
\centering
\small
\renewcommand{\arraystretch}{1.1} 
\setlength{\tabcolsep}{3pt}
\resizebox{0.47\textwidth}{!}{ 
\begin{tabular}{lccccc}
\toprule
\multirow{2}{*}{\bf Benchmarks} & \textbf{Cross-Modal} & \textbf{Cross-Modal} & \textbf{Contain} & \textbf{Contain} & \textbf{Domain} \\
          & \textbf{2-Hop?} & \textbf{3-Hop?}  & \textbf{Table?}  & \textbf{Image?}  & \textbf{Specific?} \\ 
\midrule
\textbf{ManyModalQA} & {\color{red}\LARGE\ding{55}} & {\color{red}\LARGE\ding{55}} & {\color{ao(english)}\LARGE\ding{51}} & {\color{ao(english)}\LARGE\ding{51}} & {\color{red}\LARGE\ding{55}} \\
\textbf{CT2C-QA}      & {\color{red}\LARGE\ding{55}} & {\color{red}\LARGE\ding{55}} & {\color{ao(english)}\LARGE\ding{51}} & {\color{ao(english)}\LARGE\ding{51}} & {\color{ao(english)}\LARGE\ding{51}} \\
\textbf{MME-Finance}      & {\color{red}\LARGE\ding{55}} & {\color{red}\LARGE\ding{55}} & {\color{ao(english)}\LARGE\ding{51}} & {\color{ao(english)}\LARGE\ding{51}} & {\color{ao(english)}\LARGE\ding{51}} \\
\textbf{WebQA}        & {\color{ao(english)}\LARGE\ding{51}} & {\color{red}\LARGE\ding{55}} & {\color{red}\LARGE\ding{55}} & {\color{ao(english)}\LARGE\ding{51}} & {\color{red}\LARGE\ding{55}} \\
\textbf{MuMuQA}       & {\color{ao(english)}\LARGE\ding{51}} & {\color{red}\LARGE\ding{55}} & {\color{red}\LARGE\ding{55}} & {\color{ao(english)}\LARGE\ding{51}} & {\color{ao(english)}\LARGE\ding{51}} \\
\textbf{FinQA}        & {\color{ao(english)}\LARGE\ding{51}} & {\color{red}\LARGE\ding{55}} & {\color{ao(english)}\LARGE\ding{51}} & {\color{red}\LARGE\ding{55}} & {\color{ao(english)}\LARGE\ding{51}} \\
\textbf{TAT-QA}       & {\color{ao(english)}\LARGE\ding{51}} & {\color{red}\LARGE\ding{55}} & {\color{ao(english)}\LARGE\ding{51}} & {\color{red}\LARGE\ding{55}} & {\color{ao(english)}\LARGE\ding{51}} \\
\textbf{HybridQA}     & {\color{ao(english)}\LARGE\ding{51}} & {\color{red}\LARGE\ding{55}} & {\color{ao(english)}\LARGE\ding{51}} & {\color{red}\LARGE\ding{55}} & {\color{red}\LARGE\ding{55}} \\
\textbf{OTT-QA}       & {\color{ao(english)}\LARGE\ding{51}} & {\color{red}\LARGE\ding{55}} & {\color{ao(english)}\LARGE\ding{51}} & {\color{red}\LARGE\ding{55}} & {\color{red}\LARGE\ding{55}} \\
\textbf{TANQ}         & {\color{ao(english)}\LARGE\ding{51}} & {\color{red}\LARGE\ding{55}} & {\color{ao(english)}\LARGE\ding{51}} & {\color{red}\LARGE\ding{55}} & {\color{red}\LARGE\ding{55}} \\
\textbf{MMQA}         & {\color{ao(english)}\LARGE\ding{51}} & {\LARGE\notdingcheck} & {\color{ao(english)}\LARGE\ding{51}} & {\color{ao(english)}\LARGE\ding{51}} & {\color{red}\LARGE\ding{55}} \\
\midrule
\textbf{\ours{} (Ours)} & {\color{ao(english)}\LARGE\ding{51}} & {\color{ao(english)}\LARGE\ding{51}} & {\color{ao(english)}\LARGE\ding{51}} & {\color{ao(english)}\LARGE\ding{51}} & {\color{ao(english)}\LARGE\ding{51}} \\
\bottomrule
\end{tabular}
}
\caption{Comparison of datasets based on cross-modal reasoning and modality coverage. MMQA’s \protect\notdingcheck\ shows that although it includes cross-modal three-hop reasoning, such instances constitute only 0.8\% of the dataset.}
\label{tab:related_datasets}
\end{table}

\section{Related Work}
\subsection{Cross-Modal Multi-Hop Reasoning}
Benchmarking cross-modal multi-hop reasoning has received considerable attention.
Efforts include WebQA \cite{chang2022webqa} and MuMuQA \cite{reddy2022mumuqa}---for two-hop cross-modal reasoning with text and images---as well as HybridQA \cite{chen2020hybridqa}, OTT-QA \cite{chen2020open}, FinQA \cite{chen2021finqa}, TAT-QA \cite{zhu-etal-2021-tat}, and TANQ \cite{akhtar2024tanq}, which include text and tables.
However, all these datasets are limited to \textbf{only two modalities}, making them inadequate for evaluating more complex cases.

Meanwhile, datasets like ManyModalQA \cite{hannan2020manymodalqa}, CT2C-QA \cite{zhao2024ct2c}, and MME-Finance \cite{gan2024mmefinancemultimodalfinancebenchmark} incorporate three modalities but lack an inherent focus on cross-modal multi-hop reasoning.

MMQA \cite{talmor2021multimodalqa}, in contrast, deals with \textbf{three modalities}---text, tables, and images---and requires \textbf{three-hop reasoning}, setting it apart from others.
It has served as the de facto standard for evaluating related methods \cite{rajabzadeh2023multimodalmultihopquestionanswering, yu-etal-2023-unified, luo-etal-2023-unifying, zhang2024entailment, abaskohi2024fm2ds}.
MMCoQA \cite{li2022mmcoqa} and MMCV \cite{wang2024piecing} have been developed as extensions of MMQA.

The characteristics of the benchmarks are summarized in Table \ref{tab:related_datasets}.
\ours{} is crafted to address the limitations of previous ones, particularly MMQA.

\subsection{Limitations of MMQA}
We briefly revisit the drawbacks MMQA to emphasize the need for a new, robust benchmark for cross-modal multi-hop reasoning.

\paragraph{Data Contamination}
As MMQA is constructed from Wikipedia, it is vulnerable to data contamination.
That is, the model being tested may already possess internalized knowledge of certain facts, reducing its dependence on the dataset's provided input.
In particular, Table \ref{tab:mmqa_gpt4o_performance} shows that GPT-4o can achieve reasonable performance on the most challenging part of MMQA---questions intentionally tailored for requiring a combination of information from three modalities---\textit{without} relying on visual hints.\footnote{Figure \ref{fig:appendix_mmqa_comtamination_ex} presents an example of data leak in MMQA.}
This suggests that MMQA falls short of effectively measuring cross-modal multi-hop reasoning ability as it was originally intended.

\begin{table}[t]
    \centering
    \small
    \resizebox{0.49\textwidth}{!}{
    \begin{tabular}{lccc}
        \toprule
        \textbf{Dataset: MMQA} & \textbf{Image?} & \textbf{Exact Match (\%)} & \textbf{F1 Score (\%)} \\
        \midrule
        \textbf{Random Selection} & - & 0.0 & 1.2 \\
        \midrule
        \multirow{2}{*}{\bf GPT-4o} & \color{red}\ding{55} & 43.4 & 46.2 \\
                                    & \color{ao(english)}\ding{51} & 63.4 & 67.5 \\
        \bottomrule
    \end{tabular}}
    \caption{Experiments on a subset of MMQA requiring cross-modal three-hop reasoning reveal that GPT-4o performs reasonably well even without images. This implies that it already contains information derivable from input images, questioning the rigor of MMQA.  
    For more details, refer to Appendix \ref{subsec:datacontamination}.
    }
\label{tab:mmqa_gpt4o_performance}
\end{table}

\paragraph{Lack of Cross-Modal Three-Hop Cases}
Only about 0.8\% (205 instances) of the MMQA dataset consists of cross-modal three-hop reasoning,
while the majority comprises either one-hop or two-hop questions.
This scarcity restricts its effectiveness in thoroughly evaluating a model’s performance on complex reasoning tasks with intricate interactions across text, tables, and images.

\begin{figure*}[t]  
\centering
\includegraphics[width=0.95\textwidth, keepaspectratio]{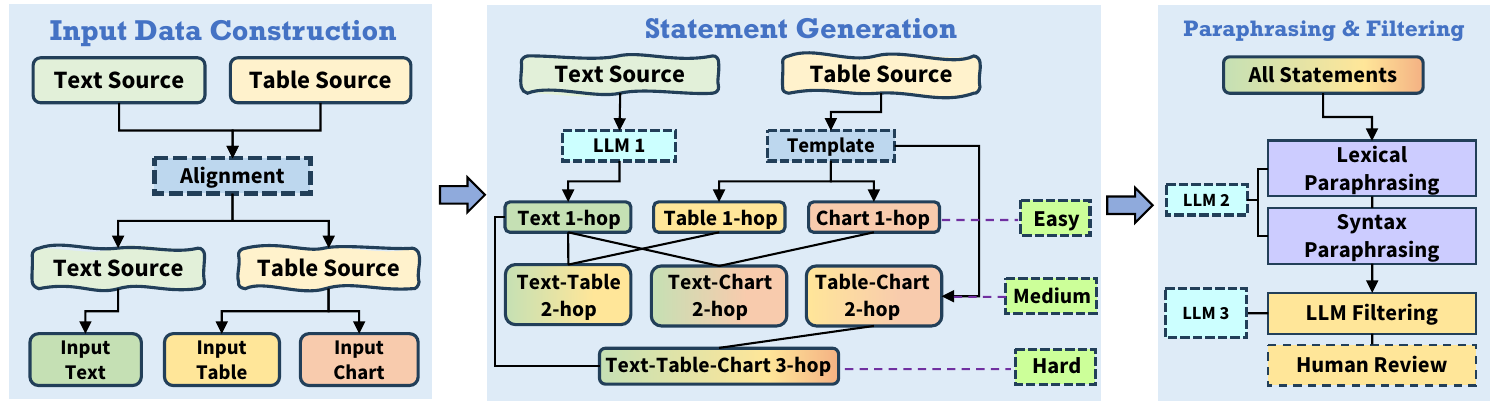}
  \caption{\textbf{\ourGen{}} is an automated and efficient framework for building cross-modal multi-hop reasoning datasets. \textbf{(1) Modality Data Construction} extracts text, table, and chart modalities from sources sharing common entities. \textbf{(2) Statement Generation} produces cross-modal one-, two-, and three-hop statements using LLM and templates. \textbf{(3) Paraphrasing \& Filtering} applies two-stage paraphrasing with LLMs, followed by LLM \& Human filtering.}
\label{fig:generation_pipeline}
\end{figure*}

\section{Proposed Benchmark: \ours{}}

We introduce \textbf{\oursfullname{} (\ours{})}, a new benchmark created to alleviate the shortcomings of MMQA and enable a more comprehensive evaluation of cross-modal multi-hop reasoning.
\ours{} includes three modalities---text, tables, and charts---and presents questions that entail selecting all correct statements from a set of three. 
The tested model must identify all true statements in the problem and is considered correct only if its final prediction is accurate.
This task inherently requires the model to address numerous subtasks---e.g., numerical calculation and chart interpretation---that require deep reasoning.

\subsection{Datset Generation Framework: \ourGen{}} 
We propose \textbf{\ourGenfullname{} (\ourGen{})}, a framework that facilitates the construction of cross-modal multi-hop reasoning datasets across various domains.
\ourGen{} distinguishes itself from other cross-modal multi-hop dataset generation methods with its highly automated and cost-effective pipeline.
Notably, while generating a single question in MMQA incurs a cost of \$0.33, our method reduces this to \$0.004 per question.
Furthermore, the framework demonstrates high adaptability to various domains and offers seamless control over difficulty levels, ranging from Easy to Hard.
In this study, we focus on the financial domain, where complex interactions among text, tables, and charts frequently occur, making it an ideal testbed for evaluating cross-modal multi-hop reasoning.
However, the proposed pipeline is also readily applicable to domains other than finance, demonstrating its flexibility.\footnote{We showcase the application of the proposed method in material science. For more details, refer to Appendix~\ref{sec:appendixmaterialqa}.}

\subsection{Procedure of \ourGen{}} 
\textbf{\ourGen{}} have three stages, as depicted in Figure \ref{fig:generation_pipeline}: (1) Input Data Construction, (2) Statement Generation, and (3) Paraphrasing \& Filtering.
We explain each step using \textbf{\ours{}} as an example.
Details of the procedure can be found in Appendix~\ref{sec:appendix_FCMU_Generation_Process}.

\paragraph{(1) Input Data Construction} 
In the first stage, we prepare and organize data for the text, table, and chart modalities. 
As the origin of information, \ourGen{} utilizes two sources: Text Source and Table Source.
For \ours{}, the Text Source consists of Annual 10-K Reports collected from the SEC EDGAR database,\footnote{\url{https://www.sec.gov/search-filings/}} while the Table Source is derived from Annual Simplified Financial Statements provided by WRDS Compustat.\footnote{\url{https://wrds-www.wharton.upenn.edu/pages/grid-items/compustat-global-wrds-basics/}}
We then filter entries that share common company entities, aligning the two sources.
Finally, we construct each data instance in \ours{}, comprising a document, a table, and a chart about three companies.
In the next step, this instance will be supplemented with three statements serving as questions about the contents generated in this stage. 
Note that the chart is created by plotting specific columns from the Table Source using custom scripts, and the columns used to create the chart are removed from the table.
We create synthetic charts but ensure diversity by incorporating various chart types and styles, as well as using different libraries (see Appendix \ref{sec:appendix_input_chart_code_generation}).

\paragraph{(2) Statement Generation} 
In the second phase, diverse forms of statements (i.e., questions) are crafted for each \ours{} instance.
We leverage GPT-4o-mini
to generate text-based one-hop statements by extracting relevant facts from the Text Source.
In addition, by leveraging various templates tailored to real-world financial scenarios—such as Trend, Ranking, Conditional Threshold, and Arithmetic—we create table-based and chart-based one-hop statements based on the Table Source.
We then combine these single-modal one-hop statements across entities to construct cross-modal two-hop statements, which are further merged to create cross-modal three-hop statements.
Each statement is categorized into Easy, Medium, or Hard based on the number of hops required for reasoning.
The complete taxonomies of statement types and templates are presented in Table \ref{tab:taboptiontypeexample} and Table \ref{tab:tabtemplate}.

\paragraph{(3) Paraphrasing \& Filtering} 
Lastly, 
we apply two stages of lexical and syntactic paraphrasing using GPT-4o to enhance the diversity of expressions in statements.
Afterward, we conduct LLM-based filtering, using Claude 3.5 Sonnet, to ensure semantic accuracy. 
By employing separate LLMs in each step, we aim to mitigate model-oriented biases.
Our framework also allows for an optional human expert review to improve their quality further. For FCMR, we use this review step to ensure high standards, especially for challenging (i.e., Hard-level) problems.
Appendix \ref{subsec:paraphrasing_quality} provides the specific details of this human annotation process. 


\subsection{Distractor Generation} 
Instead of simply adjusting numerical values to generate incorrect statements, we reflect realistic scenarios in the financial domain, where analysis of multiple companies is common, by generating distractors based on corporate entities. 
Since each statement in Easy, Medium, and Hard levels is combined with corporate entities, we generate distractors by replacing them with other companies in the same instance.

\subsection{Multiple-Choice Design} 
Previous research on cross-modal multi-hop reasoning has often employed descriptive or short-answer formats, evaluated with metrics like F1 or Exact Match. 
These approaches might result in inaccurate evaluations by penalizing semantically appropriate answers that slightly deviate in form. 
To address this, we adopt a multiple-choice format with three statements. 
Recent work \cite{pang2024uncovering} argues that single-choice question formats are more suitable for model evaluation than free-form answers, supporting our decision.
In contrast to existing multimodal benchmarks \cite{li2023seedbenchbenchmarkingmultimodalllms, 10656299, pmlr-v235-ying24a, 10.5555/3666122.3666362, 10.1007/978-3-031-72658-3_13} that typically rely on a single correct answer, our setup allows for zero to three correct statements. 
This strategy enhances the complexity of the reasoning process needed to answer the problem accurately, requiring a more comprehensive synthesis of all provided statements. 
It also allows for precise evaluation of models’ cross-modal multi-hop reasoning abilities.

\begin{table}[t]
    \centering
    \small
    \begin{tabular}{lcc}
        \toprule
        \textbf{Dataset} & \textbf{WPD} & \textbf{LD} \\
        \midrule
        MRPC & 0.12 & 0.42 \\
        PAWS & 0.07 & 0.13 \\
        \cmidrule(lr){1-3}
        \ours{} (Ours) & \textbf{0.2} & \textbf{0.45} \\
        \bottomrule
    \end{tabular}
    \caption{WPD (Word Position Deviation) represents syntactic diversity, and LD (Lexical Deviation) reflects lexical diversity. Both metrics indicate that higher scores correspond to greater diversity in paraphrasing.}
    \label{tab:dataset_wpd_ld_comparison}
\end{table}

\begin{table}[t]
    \centering
    \small 
    \begin{tabular}{lcc}
        \toprule
        \textbf{Dataset: \ours{} (Hard)} & \textbf{Image?} & \textbf{Accuracy} \\ 
        \midrule
        \textbf{Random Selection} & - & 12.28 \\
        \midrule
        \multirow{2}{*}{\bf GPT-4o} & \color{red}\ding{55} & 14.71 \\
                                    & \color{ao(english)}\ding{51} & 24.37 \\
        \bottomrule
    \end{tabular}
    \caption{    
    Replication of the experiments from Table \ref{tab:mmqa_gpt4o_performance} with \ours{}. GPT-4o’s performance drops to near random selection when charts are omitted, suggesting that \ours{} is relatively robust against data contamination.}
    \label{tab:fcmu_hard_chart_performance}
\end{table}

\subsection{Data Quality Control}
To uphold high data quality, we implement multifaceted verification protocols.\footnote{Refer to Appendix \ref{sec:appendix_details_data_quality_control} for full details of our strategies.} 
Specifically, we utilize Word Position Deviation (WPD) and Lexical Deviation (LD) metrics \cite{liu-soh-2022-towards} to evaluate paraphrasing quality and compare these values with those from MRPC \cite{dolan-brockett-2005-automatically} and PAWS \cite{zhang-etal-2019-paws}.
The outcomes, presented in Table~\ref{tab:dataset_wpd_ld_comparison}, confirm the superiority of our paraphrasing method.
Furthermore, to verify that our dataset avoids the contamination issue identified in MMQA, we replicate the contamination experiment under the same conditions.
As shown in Table \ref{tab:fcmu_hard_chart_performance}, when the chart images are withheld, GPT-4o's performance approximates random guessing, alleviating the risk of data contamination in \ours{}.
We further mitigate potential biases by equalizing modality order, statement types, and answer distributions. 
Figure \ref{fig:appendix_type_pie_chart} demonstrates that our benchmark is well-balanced across various perspectives.
The final dataset includes 757 Easy, 728 Medium, and 714 Hard instances.
Hard-level problems demand precise cross-modal three-hop reasoning. While statements in Easy and Medium questions involve simpler reasoning (e.g., single-modal one-hop for Easy), resolving each instance still necessitates processing information from all three provided modalities---text, table, and chart.

\subsection{Chart Type Coverage}
To ensure sufficient diversity in chart types, FCMR includes four commonly used formats in the financial domain: line, bar, scatter, and pie charts.
These chart types reflect the typical visualizations found in real-world financial documents.
A recent analysis \cite{christensen2024data} of 23,906 quantitative infographics extracted from all 10-K reports filed with the SEC between 2003 and 2019 shows that bar charts account for 55.0\%, pie charts for 30.6\%, and line charts for 12.7\%. 
The chart types included in our FCMR dataset directly correspond to these prevalent quantitative categories, collectively covering approximately 98\% of the most frequently used quantitative visualization types in actual 10-K filings. Details on the chart generation process are provided in Appendix \ref{sec:appendix_input_chart_code_generation}.



\begin{table}[t!]
\centering
\small
\begin{adjustbox}{max width=0.47\textwidth}
\begin{tabular}{@{}lcccc@{}}
\toprule

\textbf{Metric: Accuracy (\%)} & \textbf{\begin{tabular}[c]{@{}c@{}}Easy\end{tabular}} & \textbf{\begin{tabular}[c]{@{}c@{}}Medium\end{tabular} } & \textbf{\begin{tabular}[c]{@{}c@{}}Hard\end{tabular}} & \textbf{\begin{tabular}[c]{@{}c@{}}Avg\end{tabular}} \\
\midrule
Random Selection & 12.2 & 12.91 & 12.28 & 12.46 \\ 
\midrule
\multicolumn{5}{c}{\textbf{Multimodal Large Language Models (MLLMs)}} \\ \midrule
ChartInstruct-Llama2  & 11.49 & 12.64 & 10.78 & 11.64 \\
llama3-llava-next-8b-hf  & 16.86 & 12.22 & 11.53 & 13.54 \\
MiniCPM-V-2\_6  & 16.38 & 11.68 & 13.17 & 13.74 \\
Qwen2-VL-7B-Instruct & 17.57 & 13.32 & 12.04 & 12.32 \\
Llama 3.2 90B-Vision & 42.47 & 21.60 & 13.73 & 25.94  \\  \midrule
GPT-4o mini & 49.14 & 21.98 & 13.03 & 28.05 \\
Gemini 1.5 Flash & 57.33 & 26.65 & 13.43 & 32.80 \\
Gemini 1.5 Pro & 63.01 & 31.18 & 22.27 & 38.82 \\
GPT-4o & 64.20 & 43.70 & 24.37 & 44.09 \\
Claude 3.5 Sonnet & \textbf{75.43} & \textbf{50.82} & \textbf{30.39} & \textbf{52.21} \\ \midrule
\multicolumn{5}{c}{\textbf{Large Language Models (LLMs) with Deplot}} \\ \midrule
Qwen2-7B-Instruct & 21.66 & 11.95 & 14.01 & 15.87 \\
Llama 3.1 8B-Instruct & 30.91 & 13.05 & 10.36 & 18.11 \\
Llama 3.1 70B-Instruct & 46.37 & 17.86 & 14.01 & 26.08 \\
Llama 3.2 90B-Vision & 50.20 & 22.39 & 11.90 & 28.16  \\  \midrule
GPT-4o mini & 57.60 & 26.51 & 12.61 & 32.24 \\
GPT-4o & \textbf{68.69} & \textbf{49.18} & 32.91 & \textbf{50.26} \\
Claude 3.5 Sonnet & 66.84 & 46.15 & \textbf{36.13} & 49.71 \\
\bottomrule
\end{tabular}
\end{adjustbox}
\caption{Performance of MLLMs and LLMs on \ours{}. For LLMs, charts are converted into tables using Deplot. The best performance at each difficulty level and category is highlighted in \textbf{bold}.}
\label{tab:overall_results}
\end{table}

\section{Experiments}
\subsection{Experimental Setup}
We evaluate a wide range of MLLMs and LLMs on \ours{} under a zero-shot CoT setting, where no task-specific tuning or demonstration is provided. 
This setting reflects common use cases for MLLMs, ensuring an unbiased model evaluation and capturing overall performance trends.
All models are prompted with the same template in Figure \ref{fig:figappendixprompt0shot}. 
Tables are represented in JSON format.
For proprietary models, we employ the Claude version \textit{claude-3-5-sonnet-20241022}, the GPT-4o version \textit{gpt-4o-2024-08-06.}, and Gemini version \textit{gemini-1.5-pro-002}.
We also test several open-source models: Llama variants \cite{touvron2023llamaopenefficientfoundation}, Qwen \cite{yang2024qwen2technicalreport}, MiniCPM \cite{yao2024minicpmvgpt4vlevelmllm}, Llava \cite{Liu_2024_CVPR}, and ChartInstruct \cite{masry-etal-2024-chartinstruct}.

\subsection{Main Results}

\paragraph{Performance of MLLMs} 
Table \ref{tab:overall_results} reports the performance of various MLLM across different levels.
Most open-source models perform just above random chance at the Easy level, focused on single-modal, one-hop reasoning, confirming \ours{} as a challenging benchmark.
Proprietary models perform better, demonstrating a remarkable gap in reasoning ability.
However, at the Hard level, which necessitates full cross-modal three-hop integration, even sophisticated models, including Claude 3.5 Sonnet, achieve only around 30\%.\footnote{Recently, deep reasoning models, e.g., OpenAI’s o1 \cite{openai2024o1}, have emerged. While they are not directly related to multimodal functionality, we also evaluate them at the Hard level. As a result, we confirm that o1 achieves 43\% accuracy, indicating they are still far from perfect (refer to Table \ref{tab:o1_like_hard_sample_100}).}
This result underscores \ours{}’s challenging nature and the need for developing more advanced reasoning strategies.


\begin{figure*}[t]
\centering
\resizebox{0.95\linewidth}{!}{
\begin{subfigure}[t]{0.27\textwidth}
    \centering
    \includegraphics[width=\linewidth, keepaspectratio]{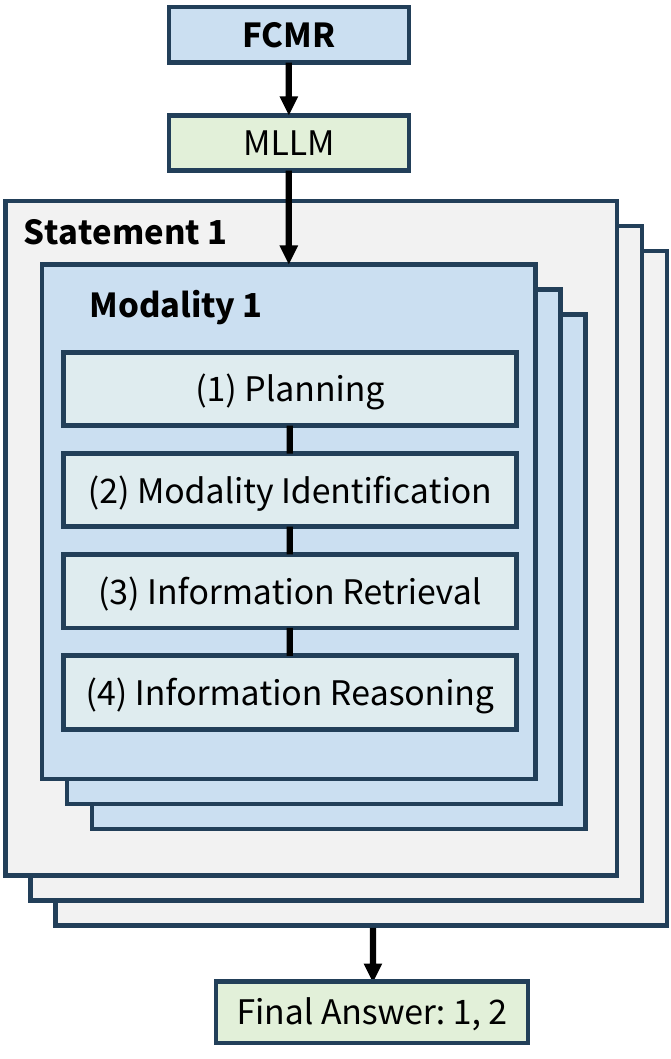}
    \caption{Fine-grained reasoning stages.}
\end{subfigure}%
\hfill
\begin{subfigure}[t]{0.7\textwidth}
    \centering
    \includegraphics[width=\linewidth, keepaspectratio]{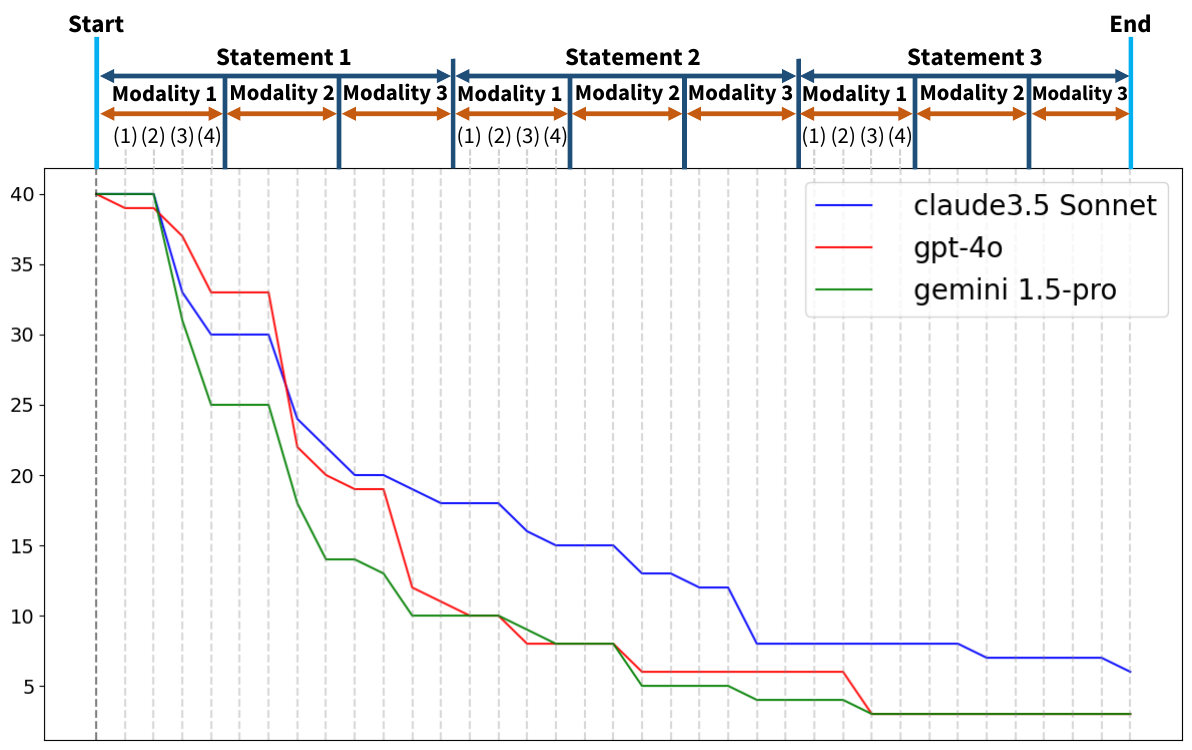}
    \caption{The number of successful inferences after each fine-grained reasoning stage.}
\end{subfigure}}
\caption{Fine-grained stage-based analysis of three advanced MLLMs. This unique strategy discovers several intriguing findings, including that the models fail most often at (3) Information Retrieval.}
\label{fig:combined_figure}
\end{figure*}

\paragraph{Performance of (M)LLMs + Deplot}
For image-blind standard large language models (LLMs), we use Deplot \cite{liu-etal-2023-deplot} to convert charts into tables, ensuring that all models can access chart information.
We also explore applying the same heuristic to MLLMs, as the literature suggests that MLLMs tend to rely more on textual clues than visual ones \cite{rahmanzadehgervi2024vision}.

Experimental results indicate that open-source models with fewer than 8B parameters continue to perform comparably to random selection for tasks at the Medium and Hard levels.
However, for the Easy category, they demonstrate superior performance compared to similarly sized MLLMs.
Surprisingly, even advanced MLLMs such as GPT-4o and Claude 3.5 Sonnet achieve performance gains in certain cases, suggesting that their visual interpretation capabilities are still not perfect.
In Section \ref{sec:chart_interpretability}, we dive deeper into this phenomenon.



\begin{figure}[t]  
\centering
\includegraphics[width=0.95\columnwidth, keepaspectratio]{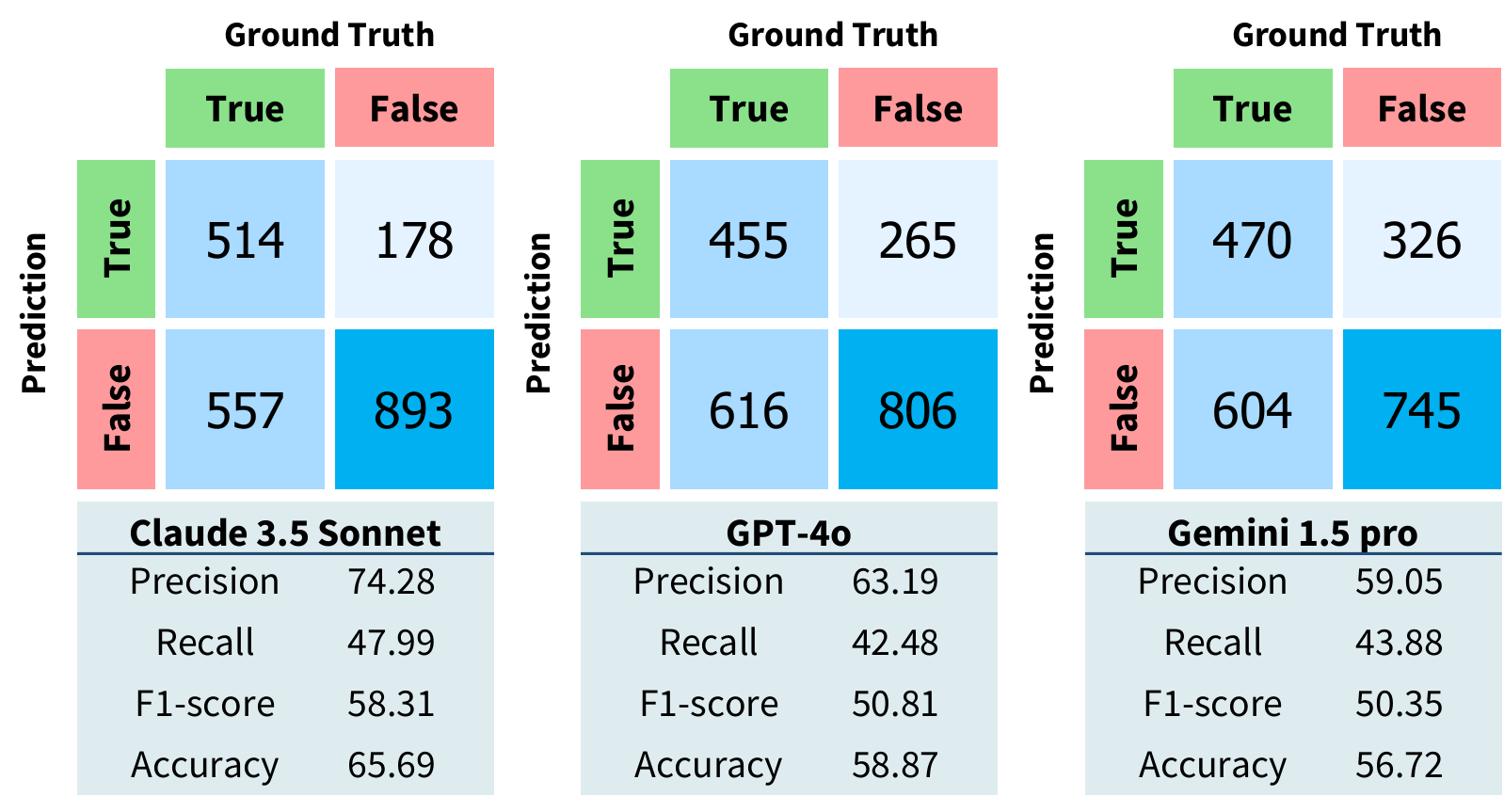}
  \caption{Confusion matrices for three advanced MLLMs, with metrics in percentages (\%).}
\label{tab:confusion_matrix}
\end{figure}


\section{Analysis}

In this section, we analyze the inner workings of closed-source MLLMs—GPT-4o, Claude 3.5 Sonnet, and Gemini 1.5 Pro—which outperform smaller open-source models.
We focus on their performance at the Hard level, as this subset presents the most challenging questions for the models.


\subsection{Statement-Level Analysis}
While each problem in \ours{} requires models to draw an overall conclusion on three statements, their partial solutions for each statement can provide insight into how well each model handles diverse cases.
We gather statistics on each model’s predictions for every statement and construction confusion matrices for analysis.
We have 714 $\times$ 3 = 2,142 statements with gold-standard answers, evenly split into true (1,071) and false (1,071) ones. 
Each model’s prediction is annotated for these statements, forming the matrices in Figure \ref{tab:confusion_matrix}.

While all three models show limitations in precision, recall, and F1-score, Claude 3.5 Sonnet achieves comparatively better performance.
With a high precision of 74.27, Claude effectively minimizes false positives, reflecting an ability to accurately classify positive cases.
However, its recall remains limited to 47.99, indicating a reduced capacity to capture all true positives. 
Despite this trade-off, Claude achieves the highest accuracy at 65.69 and an F1-score of 58.31, outperforming the other two models, GPT-4o and Gemini 1.5 Pro.

Moreover, all three models display a notable tendency to adopt a conservative decision-making strategy by defaulting to \textit{false} in cases of uncertainty or low confidence. 
This behavior reflects a low-risk approach aimed at reducing false positive classifications, even if it results in a lower recall.

\subsection{Stage-Based Analysis}
\label{stage-based-analysis}
In the main experiments, we observed advanced MLLMs follow a similar sequence of reasoning steps to solve problems in \ours{}. 
Based on this, we define four fine-grained reasoning steps to identify where errors commonly occur. 
The four stages are specified as:
\textbf{(1) Planning:} identifying the required values,
\textbf{(2) Modality Identification:} recognizing which modality contains these values,
\textbf{(3) Information Retrieval:} extracting relevant information from the identified modality, and
\textbf{(4) Information Reasoning:} reasoning over the extracted information under the given conditions.
Each instance includes three statements, each requiring the four-step process across all three modalities.
Models must execute steps (1)-(4) three times per statement and repeat this for all three statements before answering (see Figure~\ref{fig:combined_figure} (a)).

We manually monitor MLLMs’ inference trajectories on 40 given samples.
After each fine-grained stage, we record the number of problems successfully processed by the models, forming a success history diagram in Figure~\ref{fig:combined_figure} (b).
The visualization reveals an intriguing pattern: for most samples, MLLMs fail at some stage before completing the reasoning steps for the first statement.
Specifically, we observe a notable performance drop at the [Statement 1, Modality 2, (3)-(4)] stage.
This suggests that while the models handle the first modality relatively well, they struggle considerably when they encounter a second modality.
Interestingly, GPT-4o outperforms Claude 3.5 Sonnet in the first modality reasoning step of the first statement, but Claude surpasses GPT-4o starting from the second modality phase.
As the models progress to the second statement, Claude’s performance diverges further from the others, showcasing more robust and sustained reasoning capabilities.\footnote{Fine-grained model answer examples are in Figure \ref{fig:appendix_reasoning_stage_example}.}

\begin{table}[t]
    \centering
    \footnotesize
    \setlength{\tabcolsep}{0.3em}
    \begin{tabular}{lccccc}
        \toprule
        \textbf{Level / Modality} & \textbf{Text} & \textbf{Table} & \textbf{Chart} & \textbf{Total} \\
        \midrule
        \textbf{Easy} & 1 (4\%) & 5 (21\%) & 18 (75\%) & 24 \\
        \textbf{Medium} & 5 (16\%) & 6 (19\%) & 20 (65\%) & 31 \\
        \textbf{Hard} & 6 (14\%) & 13 (32\%) & 22 (54\%) & 41 \\
        \bottomrule
    \end{tabular}
    \caption{Error counts and proportions by modality for Claude 3.5 Sonnet across 90 statements per level.
    }
    \label{tab:acc_by_modal}
\end{table}

We further explore model failures by identifying which of the four reasoning steps (1)-(4) these failures occur in, regardless of statement and modality. 
As illustrated in Figure \ref{fig:append_reason_infer_fail}, the most common cause of failure across MLLMs occurs at step (3) Information Retrieval---failing to extract the required information from the identified modality.
The second most frequent failures arise at stage (4), Information Reasoning, where models struggle to correctly apply the retrieved information to the given conditions.
Notably, Gemini 1.5 Pro exhibits a higher proportion of failures at step (4), meaning that even when it successfully retrieves information, it has difficulty reasoning over it.\footnote{Examples are shown in Figure \ref{fig:appendix_reasoning_stage_error_example} and Figure \ref{fig:appendix_reasoning_stage_error_example2}.}

While Claude and GPT make no modality identification errors (stage (2)), Gemini 1.5 Pro occasionally misidentifies modalities, such as confusing chart values with table values. Gemini 1.5 Pro has no failures at the Planning stage (stage (1)). 
In contrast, GPT and Claude sometimes skip planning for the third modality after successfully handling the first two, leading to task failure. 
This shows that while all models struggle at later reasoning steps, GPT and Claude particularly struggle to maintain a consistent strategy across modalities.


\begin{table}[t!]
\begin{center}
\footnotesize
\setlength{\tabcolsep}{0.5em}
\begin{tabular}{lccccc}
\toprule[1pt]
\textbf{Level / Type} & \textbf{Line} & \textbf{Bar} & \textbf{Scatter} & \textbf{Pie} & \textbf{Total} \\
\midrule
\textbf{Easy} & 74.89  & 78.60 & 71.01  & 84.31 & 75.43  \\
\textbf{Medium} & 52.70  & 50.00 & 49.79 & - & 50.82  \\
\textbf{Hard} & 39.22  & 29.20  & 23.44  & -  & 30.39 \\
\bottomrule[1pt]
\end{tabular}
\end{center}
\caption{
Accuracy by chart type, based on Claude 3.5 Sonnet. 
All values are presented as percentages (\%).
Pie charts are only used in the Ranking option type of the Easy difficulty, as they are unsuitable for the Trend option type.}
\label{tab:tabcharttype}
\end{table}

\section{Case Study}

Given Claude 3.5 Sonnet’s effectiveness for \ours{}, we conduct case studies to derive insights for enhancing cross-modal multi-hop reasoning.

\subsection{Error Rate by Modality}
Table \ref{tab:acc_by_modal} displays the numbers and proportions of statements Claude fails to interpret correctly, based on random 90 statements for each difficulty level.\footnote{Errors unrelated to modality, such as misinterpreting conditions, are excluded from this analysis.}
At the Easy level, Claude often struggles with analyzing charts, performing noticeably worse compared to its handling of text and tables. 
This disparity indicates that MLLMs exhibit weaker capabilities in interpreting charts than in processing textual or tabular data (see Appendix \ref{sec:chart_interpretability} for details).
However, as the difficulty level increases, errors in text and tables become more prominent. 
This shift is likely attributed to the increased complexity of reasoning chains, which raises the likelihood of errors in processing text and tables.

\subsection{Chart Interpretability} \label{sec:chart_interpretability}
Building on previous findings that even Claude struggles with chart interpretation, we analyze the specific conditions that pose the greatest challenges.
Table \ref{tab:tabcharttype} shows that among line, bar, scatter, and pie charts, scatter plots are the most challenging due to their less structured representations. 
In contrast, MLLMs find it easier to identify trends in line or bar charts, which provide clearer patterns. 
Ranking tasks appear simpler than trend analysis, as they involve identifying extremes, whereas trend detection demands more advanced inference.


\begin{figure}[t]  
\centering
\includegraphics[width=0.95\columnwidth, keepaspectratio]{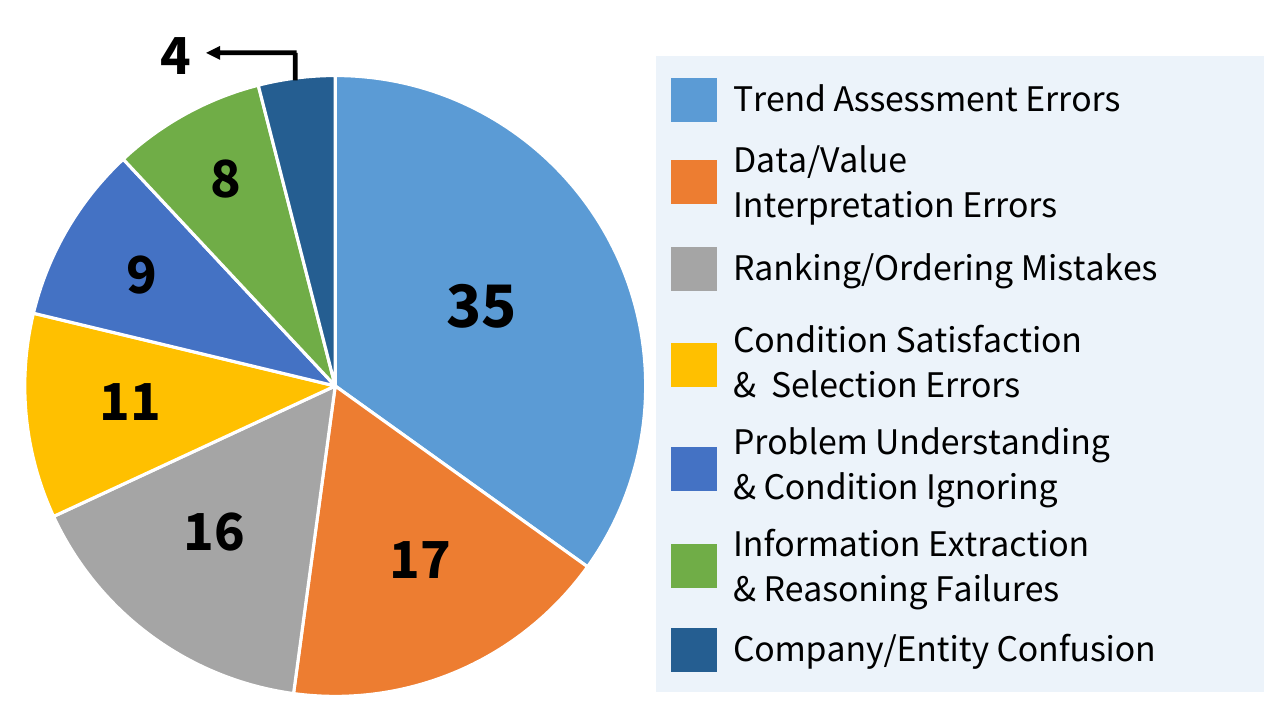}
  \caption{Error categories and their distributions across 100 manually labeled error cases for Claude 3.5 Sonnet.}
\label{fig:error_dist}
\end{figure}

\subsection{Error Classification and Inspection}
\label{subsec:Error Classification and Inspection}
To manually inspect Claude’s working patterns, we examine 100 error cases where it was unable to provide accurate answers.
The distribution of these errors is visualized in Figure \ref{fig:error_dist}. 
Real error cases can be found in Figure \ref{fig:case_exam} and Appendix \ref{sec:appendix_case_study}.



The most common issue involves misinterpreting trends in charts (35 cases).
The second most frequent error type (16 samples) pertains to the misidentification of top-ranked entities or overall rankings.
The model fails in 11 cases to correctly identify entities that meet the given conditions or misapplies the conditions.
17 errors arise from misinterpreting data, including sums, negative values, subtle differences, and column confusion. 
4 involve conflating company or entity identities. 
The model also makes 8 errors in information extraction and reasoning, such as misreading facts or drawing unjustified conclusions. 
Lastly, 9 cases stem from misinterpreting instructions, ignoring required modalities, or making illogical inferences.

These results highlight that successful cross-modal multi-hop reasoning demands both strong reasoning abilities and effective interpretative skills for each modality.
Therefore, enhancing a model’s capacity to interpret individual modalities is essential for enabling effective multi-hop reasoning.

\begin{figure}[t]  
\centering
\includegraphics[width=1\columnwidth, keepaspectratio]{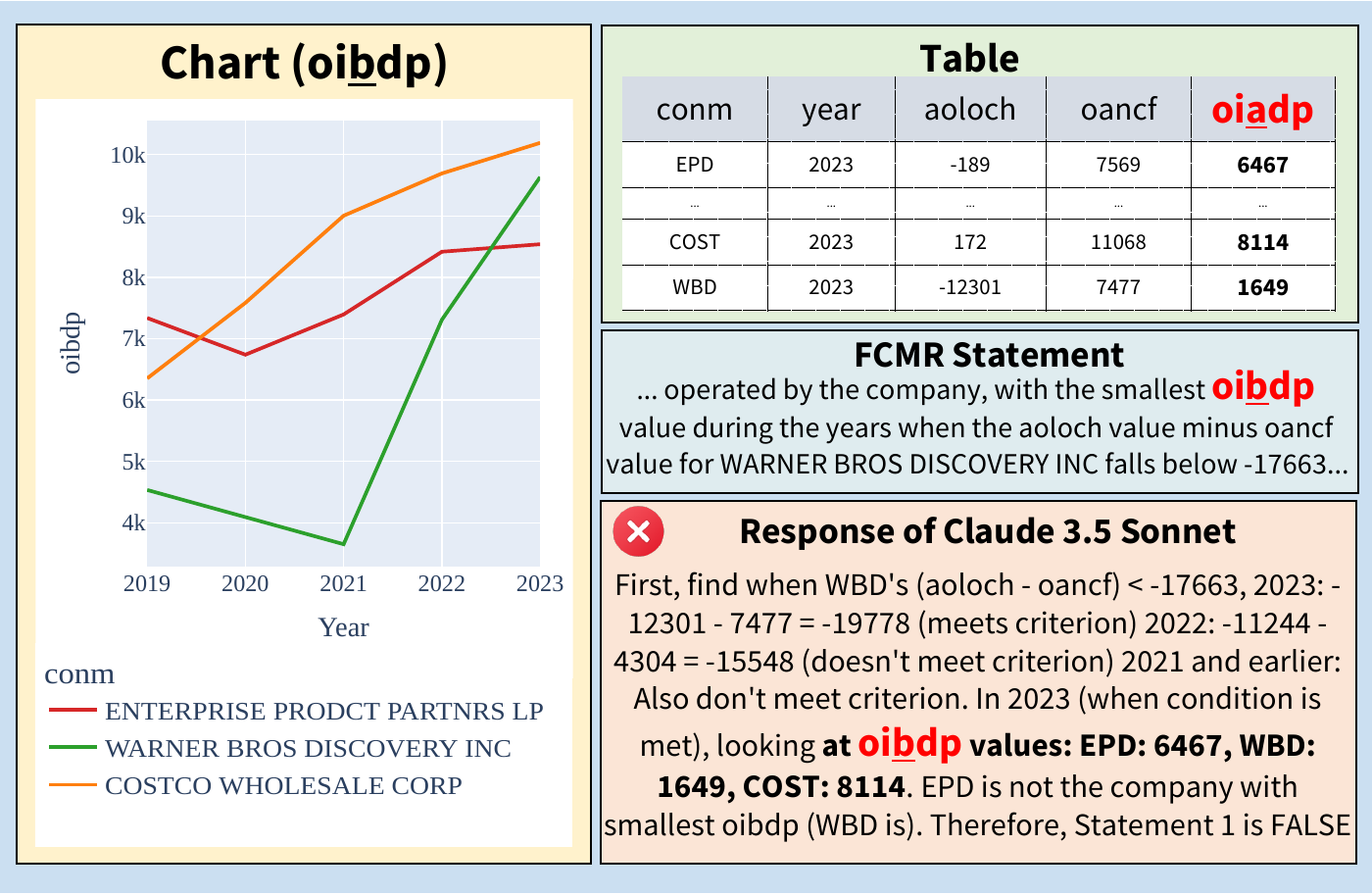}
  \caption{A case study on Claude’s errors: The condition, ``\textit{the years when the aoloch value minus the oancf value for WARNER BROS DISCOVERY INC falls below -17663,}'' is satisfied for 2023. This task thus requires finding a company with a small oibdp value for 2023. Although the \textcolor{red}{oi\underline{\textbf{b}}dp} value is in the chart, not the table, Claude ignores the chart and incorrectly substitutes the \textcolor{red}{oi\underline{\textbf{a}}dp} value, likely due to the similar column names.}
\label{fig:case_exam}
\end{figure}

\subsection{Preliminary Task Optimization}

To explore ways to enhance MLLM performance on FCMR, we conduct preliminary experiments in modality integration and prompt optimization.
Table \ref{tab:solution_hard_sample_100} presents the results of applying three techniques---Modality Integration, 4-Stage Reasoning, and Self-Refine---to 100 Hard samples.\footnote{
\textbf{Modality Integration:} convert all forms of knowledge into text representations, employing chart captioning and table linearization \cite{yu2023unifiedlanguagerepresentationquestion, luo-etal-2023-unifying, 10535103}.
\textbf{4-Stage Reasoning:} derives insights from Section \ref{stage-based-analysis} and explicitly mentions four reasoning stages in the prompt.
\textbf{Self-Refine:} instructs the model to iteratively refine its own answer \cite{madaan2023selfrefineiterativerefinementselffeedback}. 
The prompts for each method are shown in Figure \ref{fig:modality_integ_prompt_figure}, Figure \ref{fig:four_stage_reasoning_figure}, and Figure \ref{fig:self_refine_prompt_figure}, respectively.}
The results confirm each method’s contribution to improved accuracy, verifying their effectiveness.
This method also mitigates failure rates in both the Information Retrieval and Reasoning stages mentioned in Section \ref{stage-based-analysis} (see Figure \ref{fig:appendix_prompt_optim_stage_based_analysis_figure}).
However, considering that the combination of all three methods saturates at 46\% accuracy, future work is encouraged to develop dedicated approaches for FCMR.

\begin{table}[t!]
\centering
\small
\resizebox{\columnwidth}{!}{%
\begin{tabular}{@{}lc@{}}
\toprule
\textbf{Method (Claude 3.5 Sonnet; tested on 100 Hard cases)} & \textbf{Acc. (\%)} \\
\midrule
Zero-Shot COT (Baseline) & 32 \\
Modality Integration & 39 \\
Modality Integration + 4-Stage Reasoning & 42 \\
Modality Integration + 4-Stage Reasoning + Self-Refine & 46 \\
\bottomrule
\end{tabular}%
}
\caption{Preliminary task optimization results.}
\label{tab:solution_hard_sample_100}
\end{table}

\section{Conclusion}
We introduce \ours, a new benchmark designed to evaluate the cross-modal multi-hop reasoning ability of MLLMs. 
We evaluate the performance of state-of-the-art MLLMs, revealing that current models struggle with reasoning across different modalities.
In future work, we plan to develop optimized methods to enhance performance based on the observations and analyses from this study.

\section*{Limitations}

We present several points that can serve as the foundation for improving this work and initiating future research.

\paragraph{Potential for Extension to Other Domains}
While we have conducted extensive experiments and analyses in the financial domain using \ours{}, the proposed dataset generation framework, \ourGen{}, has the potential to extend beyond the financial and material science domains, enabling the creation of datasets in fields such as law, biology, medicine, and electrical engineering. 
Future work can consider performing comprehensive performance evaluations of various models across these domains.

\paragraph{Reliance on Manual Analysis}
Part of the analysis in this work is based on the manual inspection of in-house researchers.
While this was inevitable to guarantee a high-quality, in-depth investigation, future work may concentrate more on automated analysis.
We also emphasize that, despite its cost, manual analysis is worth investigating, as demonstrated in Section \ref{stage-based-analysis} and Section \ref{subsec:Error Classification and Inspection}.



\section*{Acknowledgments}

This work was supported by the National Research Foundation of Korea(NRF) grant funded by the Korea government(MSIT) (RS-2025-00558151).
This work was supported by Institute of Information \& communications Technology Planning \& Evaluation (IITP) under the artificial intelligence semiconductor support program to nurture the best talents (IITP-(2025)-RS-2023-00253914) grant funded by the Korea government(MSIT).
This work was supported by Institute of Information \& communications Technology Planning \& Evaluation (IITP) grant funded by the Korea government(MSIT) (No.RS-2020-II201373, Artificial Intelligence Graduate School Program(Hanyang University)).

\bibliography{custom}

\clearpage
\appendix

\begin{figure*}[t]  
\centering
\includegraphics[width=0.95\textwidth, keepaspectratio]{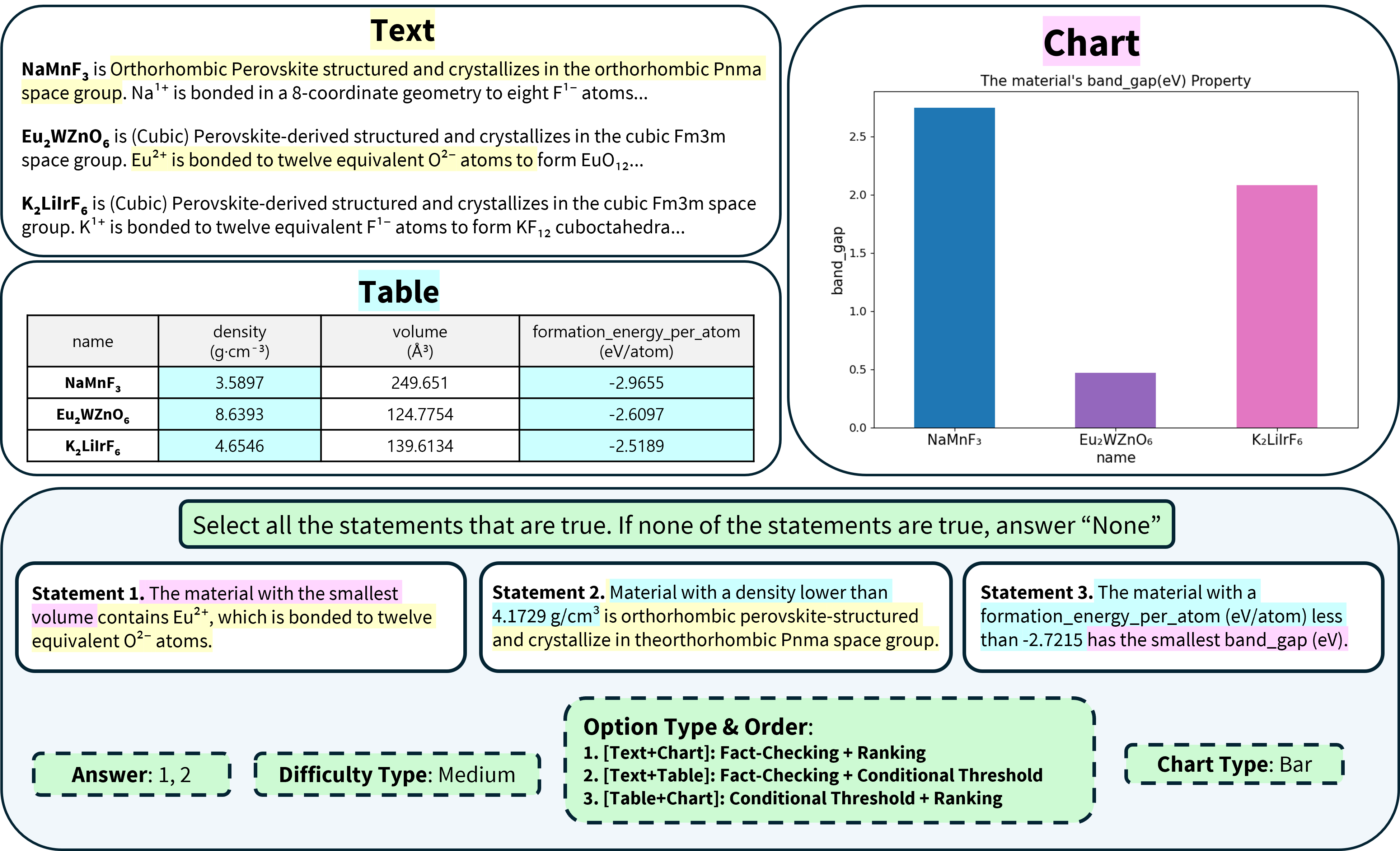}
  \caption{An example from Material-CMR.}
  \label{fig:figappendix1}
\end{figure*}

\section{Details of Material Cross-Modal Multi-Hop Reasoning (Material-CMR)} \label{sec:appendixmaterialqa}

The proposed \ourGen{} framework is easily adaptable to various domains, and as an example, we applied it to the field of Material Science to create the Material Cross-Modal Multi-Hop Reasoning (Material-CMR) dataset.
We construct a Table Source containing material properties and a Text Source describing the materials through The Materials Project. 
The Materials Project\footnote{\url{https://next-gen.materialsproject.org/}} is an initiative that accelerates materials discovery by providing researchers with computational data and tools to predict material properties, enabling more targeted and efficient experimental research.  

Specifically, we transform the entity corresponding to companies in \ours{} into materials.  
Following the same steps proposed in \ourGen{}, we build Table Sources and Text Sources for Material-CMR. 
The Table Sources include columns for material properties such as band gap, density, volume, and formation energy per atom, while the Text Sources provide descriptions of the material’s crystal structure, structural characteristics, and properties.
Using these Text Sources and Table Sources, we create datasets combining text, table, and chart input data. 
Subsequently, we generate single-modal one-hop, cross-modal two-hop, and cross-modal three-hop statements through GPT-4o-mini and templates, categorizing them by difficulty level. 
Also, A two-stage paraphrasing process is employed to maximize diversity. 
An example of the dataset instance is shown in Figure \ref{fig:figappendix1}.

\section{Details on the Procedure of \ourGen{}} \label{sec:appendix_FCMU_Generation_Process}
This section provides a comprehensive explanation of the dataset generation process for \ours{}, detailing the sources and preprocessing steps for input table, text, and chart modalities, as well as the construction of table sources, text sources, and distractors.

\subsection{Table Source}

\paragraph{WRDS Compustat: Annual Simplified Financial Statements}
Considering the practicality of cross-modal multi-hop reasoning, we utilize \textit{annual financial statements}, an essential element in real corporate analysis, as the table data source. 
Wharton Research Data Services (WRDS) Compustat\footnote{\url{https://wrds-www.wharton.upenn.edu/pages/grid-items/compustat-global-wrds-basics/}} provides various financial data of publicly traded companies in North America. 
Among them, we use the \textbf{Annual Simplified Financial Statement}, which includes key financial columns such as Company Name, Ticker Code, Year, Net Sales, and Total Assets, spanning multiple years for each company.
The Annual Simplified Financial Statement consists of 80 columns, which are broadly classified into four categories: Identifying Information, Balance Sheet Variables, Income Statement Variables, and Statement of Cash Flows Variables. The components of each category are presented in Figure \ref{fig:figcolumn}.
This Annual Simplified Financial Statement will later be used to construct the table and chart modalities.

\paragraph{Preprocessing}
The raw dataset contains a total of 80 columns. 
We standardize the unique symbol IDs to ticker codes and perform preprocessing to remove two columns that are not in millions of units to unify the units by column, leaving a total of 70 columns. 
We also use data from the most recent five years, 2019 to 2023.

\subsection{Text Source}

\paragraph{SEC EDGAR: 10-K Report}
To construct texts that are closely related to the Annual Simplified Financial Statement of company entities, we focus on corporate financial reports. 
Companies listed on the U.S. stock market are required to periodically provide financial reports to the U.S. Securities and Exchange Commission (SEC), and these reports can be publicly accessed through the Electronic Data Gathering, Analysis, and Retrieval System (EDGAR).\footnote{\url{https://www.sec.gov/search-filings}}
We use the annual disclosure report, the 10-K report, of companies to match the Annual Simplified Financial Statement. 
This 10-K report differs from the summary-style annual reports typically used in datasets such as FinQA \cite{chen2021finqa} and TAT-QA \cite {zhu-etal-2021-tat}, as it provides more in-depth financial data and disclosures.
All companies' 10-K reports are composed of a common table of contents format.

Each 10-K report includes several key items that are vital for corporate analysis. 
For instance, Item 1 provides a description of the company’s business model, its products or services, and its primary markets. 
Item 7, often referred to as the Management’s Discussion and Analysis (MD\&A), allows company executives to discuss operational results, providing insight into trends, risks, and strategies. 
In addition, Item 7A covers quantitative and qualitative disclosures regarding market risks, while Item 8 presents the audited financial statements, offering a transparent view of the company's financial health. These items, along with other sections, make the 10-K an essential document for evaluating a company’s long-term viability and strategy.

Among the various items of the 10-K report, we use ITEM 1 (Business), ITEM 2 (Legal Proceedings), ITEM 7 (Management’s Discussion and Analysis of Financial Condition and Results of Operations), ITEM 7A (Quantitative and Qualitative Disclosures about Market Risk), and ITEM 8 (Financial Statements and Supplementary Data), which are most commonly used in actual corporate analysis. Further details can be found in the document provided by the SEC\footnote{\url{https://www.sec.gov/files/reada10k.pdf}}.

This 10-K report data will later be used to construct the input text.

\paragraph{Preprocessing}
To align with the Annual Simplified Financial Statement data, we filter companies where both the Annual Simplified Financial Statements and 10-K annual reports exist, ensuring all formats of reports for the most recent five years are fully present.
Among them, we select the top 101 companies based on Net Sales in 2023.

\subsection{Table Source Construction} 
A table source serves as an intermediate bridge connecting the text, table, and chart modalities and is used as a base anchor for creating multi-hop statements. 
After sampling three companies from the Annual Simplified Financial Statement data of the 101 companies, we construct the table source by randomly sampling seven financial columns excluding company name, ticker code, and year.
One column, used for chart generation, is chosen to avoid NaN values.
The final generated table source consists of the Annual Simplified Financial Statement data of three companies, each having five years, and is composed of ten columns.

\subsection{Text Source Construction}
The 10-K reports obtained through SEC EDGAR are too lengthy to use entirely at once as input text. 
Therefore, we divide the 10-K reports of each company into chunks of three consecutive paragraphs.
Later, these chunks will be used as the input text.

\subsection{Input Data Construction: Text, Tables, and Charts}
A single data instance contains a total of three companies. 
From the table source with three companies, we select one column without NaN values as the chart column and convert the table source excluding the chart column into the input table and the chart column into the input chart.
To preserve the structural information of the table, the input table is constructed in JSON format, and to ensure data diversity, the input chart uses three different libraries and four chart types (line, bar, scatter, pie) commonly used in financial domains. 
The input text corresponds to the text sources of the three companies.
All of these processes are automated through a Python script.


\subsection{Input Chart Code Generation}
\label{sec:appendix_input_chart_code_generation}
The input chart in \ours{} consists of four types: Line, Bar, Scatter, and Pie, generated using visualization libraries such as matplotlib, seaborn, and plotly. 
To enhance chart diversity and mitigate data bias, 16 font types, including [`Arial', `Verdana', `Times New Roman', `Courier New', `Georgia', `Comic Sans MS', `Tahoma', `Cambria', `Microsoft YaHei', `Nirmala UI', `Calibri', `Consolas', `Segoe UI', `Garamond', `Century Schoolbook', `Book Antiqua'], were applied to text within the charts.
The font size for titles, labels, legends, and ticks was randomly selected within predefined minimum and maximum thresholds. 
To ensure clear visual distinction, the color palette consisted of seven colors: [`\#1f77b4', `\#ff7f0e', `\#2ca02c', `\#d62728', `\#9467bd', `\#8c564b', `\#e377c2']. 
The thickness of lines and bars was also randomly selected within predefined thresholds. 
To clearly visualize trends and rankings in charts, we introduced controlled variance in yearly data values, ensuring the design avoids cases where differentiation is visually ambiguous.

\section{Details of Data Quality Control} \label{sec:appendix_details_data_quality_control}


\subsection{Paraphrasing Quality}
\label{subsec:paraphrasing_quality}
To evaluate the quality of Lexical-Syntax 2-Stage Paraphrasing, we employed the Word Position Deviation (WPD) and Lexical Deviation (LD) metrics proposed in \cite{liu-soh-2022-towards}. 
WPD assesses the syntactic diversity of paraphrased sentences, while LD evaluates lexical diversity. 
For an objective comparison, as shown in Table \ref{tab:dataset_wpd_ld_comparison}, we benchmarked the WPD and LD metrics of \ours{} against prominent paraphrasing datasets such as MRPC \cite{dolan-brockett-2005-automatically} and PAWS \cite{zhang-etal-2019-paws}, demonstrating the superior quality of our paraphrasing.
Additionally, we validated semantic consistency using Claude 3.5 Sonnet to filter out samples where the paraphrased sentences were flagged as semantically altered.

To ensure high quality and robustness in FCMR's Hard-level instances, a specific human annotation and review process was employed.
Human reviewers initially examined 10\% of samples from all difficulty levels (Easy, Medium, and Hard).
This initial review showed that Easy and Medium samples were of high quality, so no further extensive human annotation was deemed necessary for them.
Due to their complexity, Hard-level questions underwent a more rigorous human review to ensure their quality.
Through this focused review, 22 out of 714 Hard-level samples (approximately 3\%) were identified for revision.
These revisions aimed to improve clarity, correctness, or the intended reasoning complexity.
Finally, two authors cross-checked each revision to ensure consensus and maintain high-quality standards.

\subsection{Verification of Data Contamination}
\label{subsec:datacontamination}
To ensure a fair comparison of data contamination between MMQA and FCMR under identical conditions, we evaluated instances requiring cross-modal three-hop reasoning from each dataset using the GPT-4o model under the following settings: (1) Random Selection, (2) Without Image Input, and (3) With Image Input.
For MMQA, \textit{Random Selection} in Table \ref{tab:mmqa_gpt4o_performance} involves randomly selecting a single word from the question, text, or table. 
In contrast, for FCMR, \textit{Random Selection} in Table \ref{tab:fcmu_hard_chart_performance} involves randomly selecting one of the eight possible answers, ranging from none to (1, 2, 3).


\subsection{Bias Mitigation Strategies}
Due to the design requiring reasoning over three statements, there is a potential for bias to arise from specific factors. 
To minimize bias, we implemented several strategies. 
First, we ensured a balanced distribution of modality order types to prevent bias toward specific order configurations.
Second, we adjusted the distribution of statement types to avoid overrepresentation of particular types. 
Third, we maintained an even distribution across the eight answer types to reduce bias toward any specific answer type. 
The distributions of answer type, statement type, and library type across all difficulty levels are visualized in Figure \ref{fig:appendix_type_pie_chart}.






\section{Case Study Examples}
\label{sec:appendix_case_study}
\subsection{Trend Assessment Error}
As shown in Figure \ref{fig:appendix_case_tae}, Claude struggles to identify increasing trends.
This difficulty is particularly pronounced when interpreting cumulative bar charts or charts with ranges that include negative values, where the success rate of interpretation is significantly lower.



\subsection{Ranking/Ordering Mistake}
Figure \ref{fig:appendix_case_rom} illustrates a case where the Claude model fails to accurately determine the ranking for a specific year from a chart.
While the model performs better in identifying rankings compared to recognizing increasing or decreasing trends, its success rate remains significantly lower when interpreting cumulative bar charts or charts with ranges that include negative values.

\subsection{Condition Satisfaction \& Selection Error}
The model sometimes fails to correctly identify a company or element that meets given conditions, or asserting that no such entity exists when one does.
An example is in Figure \ref{fig:appendix_case_csse}.

\subsection{Data/Value Interpretation Error}
The model occasionally fails in calculations involving addition when negative numbers are included or when the number of terms exceeds three.
Additionally, there are instances where it fails to correctly compare the magnitude of numbers.
Figure \ref{fig:appendix_case_dvie} illustrates one such case. Considering that addition and magnitude comparison are simple operations for humans, this highlights the need for improvement in the arithmetic reasoning capabilities of MLLMs.

\subsection{Company/Entity Confusion}
Errors in this category involve mixing up one company or entity with another.
Even when companies are distinguished by unique colors, labels, or legends, the model may incorrectly assign data from one company to a different one, thus undermining the validity of its reasoning and final answers.
An example is in Figure \ref{fig:appendix_case_cec}.

\subsection{Information Extraction \& Reasoning Failure}
There are the cases incorrectly extracting facts, misunderstanding textual information, drawing unjustified conclusions, or logical missteps after gathering correct details.
An example is in Figure \ref{fig:appendix_case_ierf}.

\subsection{Problem Understanding \& Condition Ignoring}
Claude sometimes makes incorrect judgments by considering only a subset of the required conditions.
This issue is particularly prominent in Hard-level tasks that require deep reasoning. An example of this case is in Figure \ref{fig:appendix_case_puci}.



\begin{table}[t!]
\centering
\footnotesize
\resizebox{\columnwidth}{!}{%
\setlength{\tabcolsep}{0.3em}
\begin{tabular}{@{}lc@{}}
\toprule[1pt]
\textbf{Model} & \textbf{Accuracy (\%)} \\
\midrule
Claude 3.5 Zero-Shot COT Baseline & 32 \\
\midrule
Mulberry-LLaVA-8B \cite{yao2024mulberryempoweringmllmo1like} & 12 \\
Virgo 72B \cite{du2025virgopreliminaryexplorationreproducing} & 19 \\
Gemini 2.0 Flash Thinking \cite{gemini2025thinking} & 39 \\
o1 \cite{openai2024o1} & 43\\
\midrule
nova-pro-v1 \cite{amazon2024nova} & 22 \\
grok-2-vision \cite{grok2vision2024} & 19\\
qwen2.5-vl-72b-instruct \cite{qwen2025qwen25technicalreport} & 19 \\
gemma-3-27b-it \cite{gemmateam2025gemma3technicalreport} & 19\\
\bottomrule[1pt]
\end{tabular}%
}
\caption{
Performance of various advanced Multimodal Large Language Models (MLLMs), including o1-like reasoning models and other recent models, on 100 randomly selected FCMR Hard-level samples.
Evaluation was limited to this 100-sample subset due to the operational costs associated with these models.
}
\label{tab:o1_like_hard_sample_100}
\end{table}


\begin{figure*}[t]  
\centering
\includegraphics[width=0.8\textwidth, keepaspectratio]{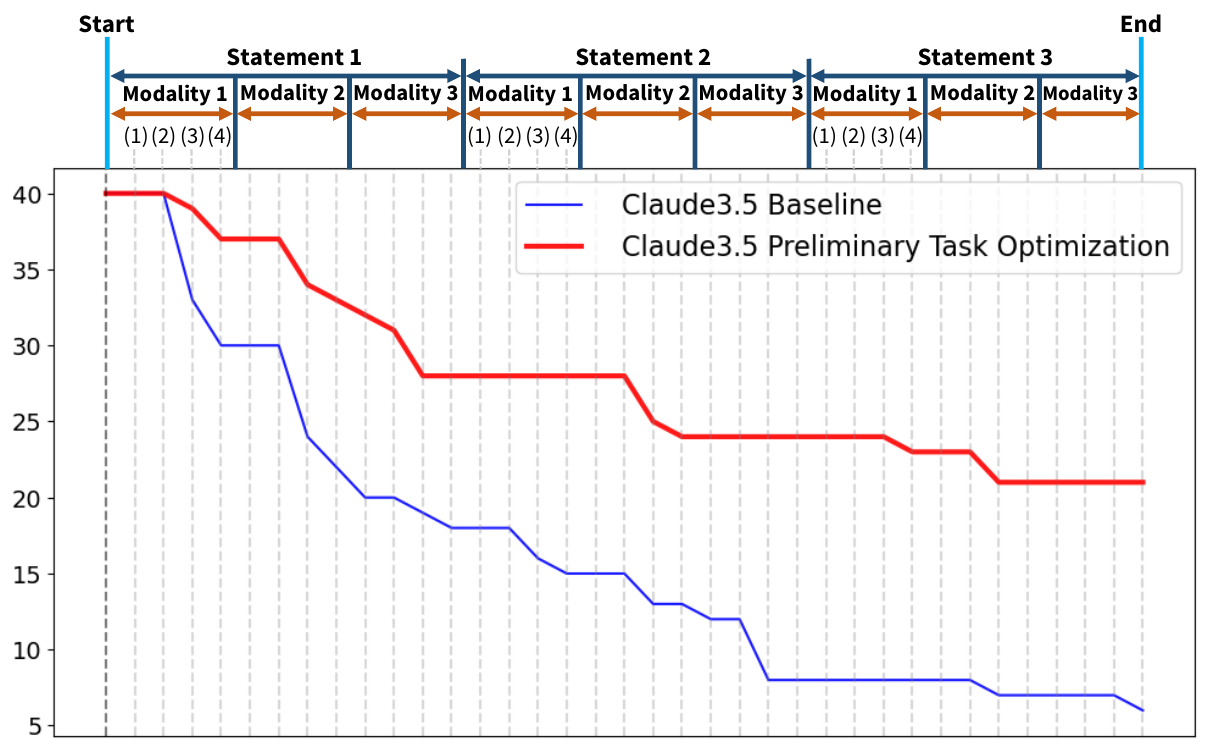}
  \caption{Fine-grained Stage-based Analysis of Preliminary Task Optimization results applying Modality Integration, 4-Stage Reasoning, and Self-Refine.}
\label{fig:appendix_prompt_optim_stage_based_analysis_figure} 
\end{figure*}

\begin{figure*}[t]  
\centering
\includegraphics[width=0.8\textwidth, keepaspectratio]{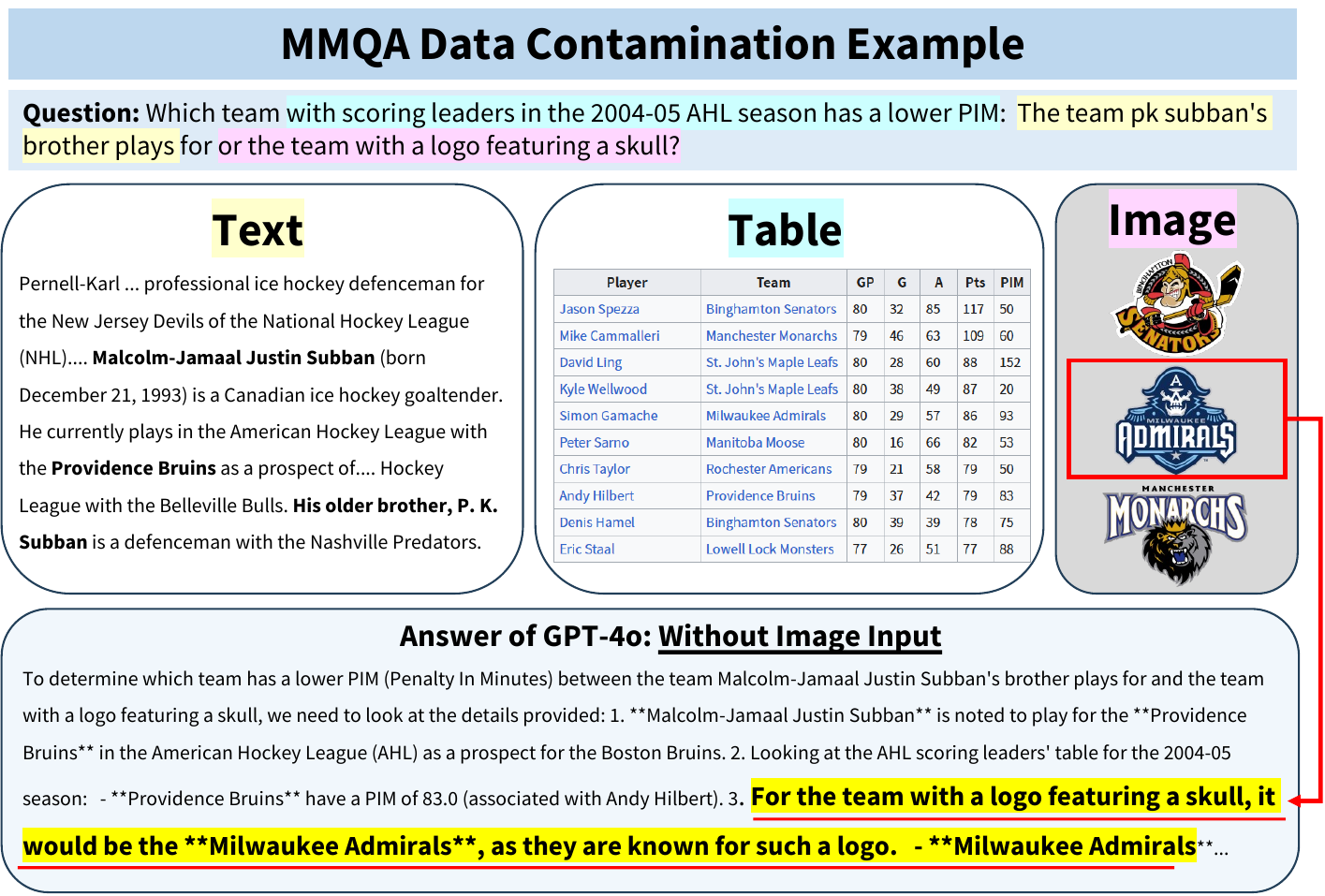}
  \caption{Example of data contamination in MMQA. Despite the requirement to infer the pink-highlighted information, \textit{"team with a logo featuring a skull"}, using image data, GPT-4o deduced the answer, \textit{"Milwaukee Admirals"}, without being provided with the image. This suggests the model leveraged internal knowledge, rather than cross-modal reasoning, highlighting a limitation in evaluating the cross-modal multi-hop reasoning capabilities of MMQA.}
\label{fig:appendix_mmqa_comtamination_ex} 
\end{figure*}

\begin{figure*}[htbp] 
    \centering
    \includegraphics[width=0.8\textwidth]{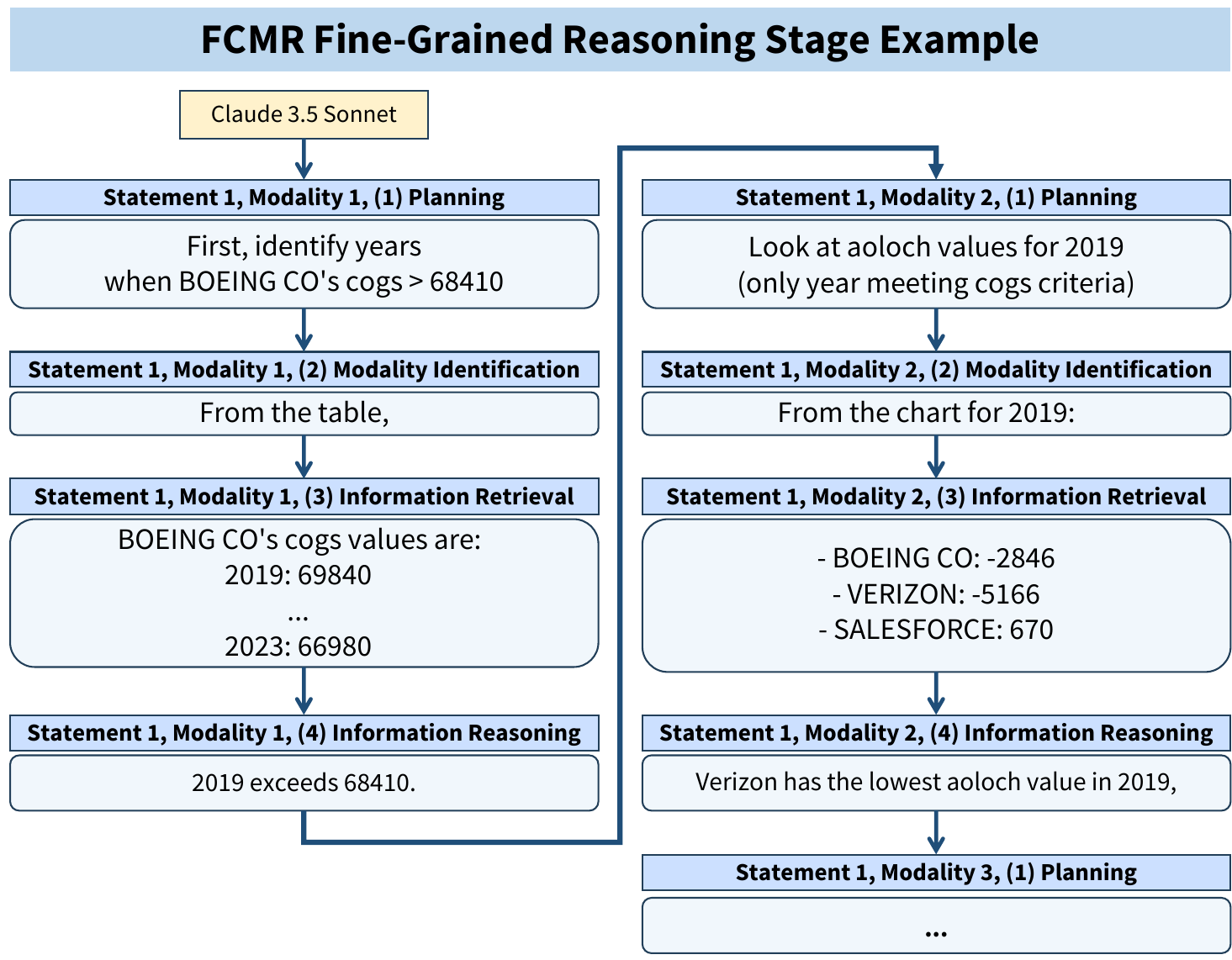} 
    \caption{An example of decomposing the reasoning process of the Claude 3.5 Sonnet's response into fine-grained, stage-based steps.}
    \label{fig:appendix_reasoning_stage_example} 
\end{figure*}

\begin{figure*}[htbp] 
    \centering
    \includegraphics[width=0.7\textwidth]{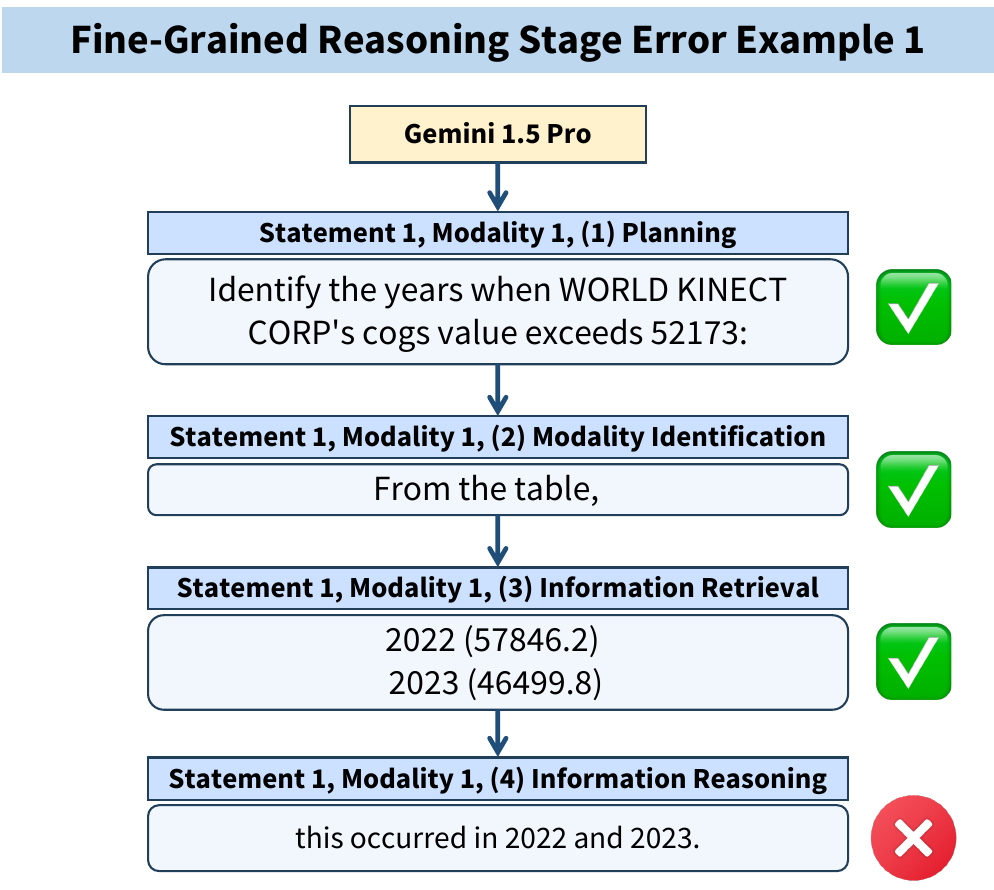}
    \caption{An example where the Gemini 1.5 Pro succeeds in (3) Information Retrieval but fails in (4) Information Reasoning. While the model successfully extracts the information that the cogs value is 57846.2 for 2022 and 46499.8 for 2023 from the table, it incorrectly reasons that the cogs values for both 2022 and 2023 are greater than 52173, resulting in a failure.}
    \label{fig:appendix_reasoning_stage_error_example} 
\end{figure*}

\begin{figure*}[htbp] 
    \centering
    \includegraphics[width=\textwidth]{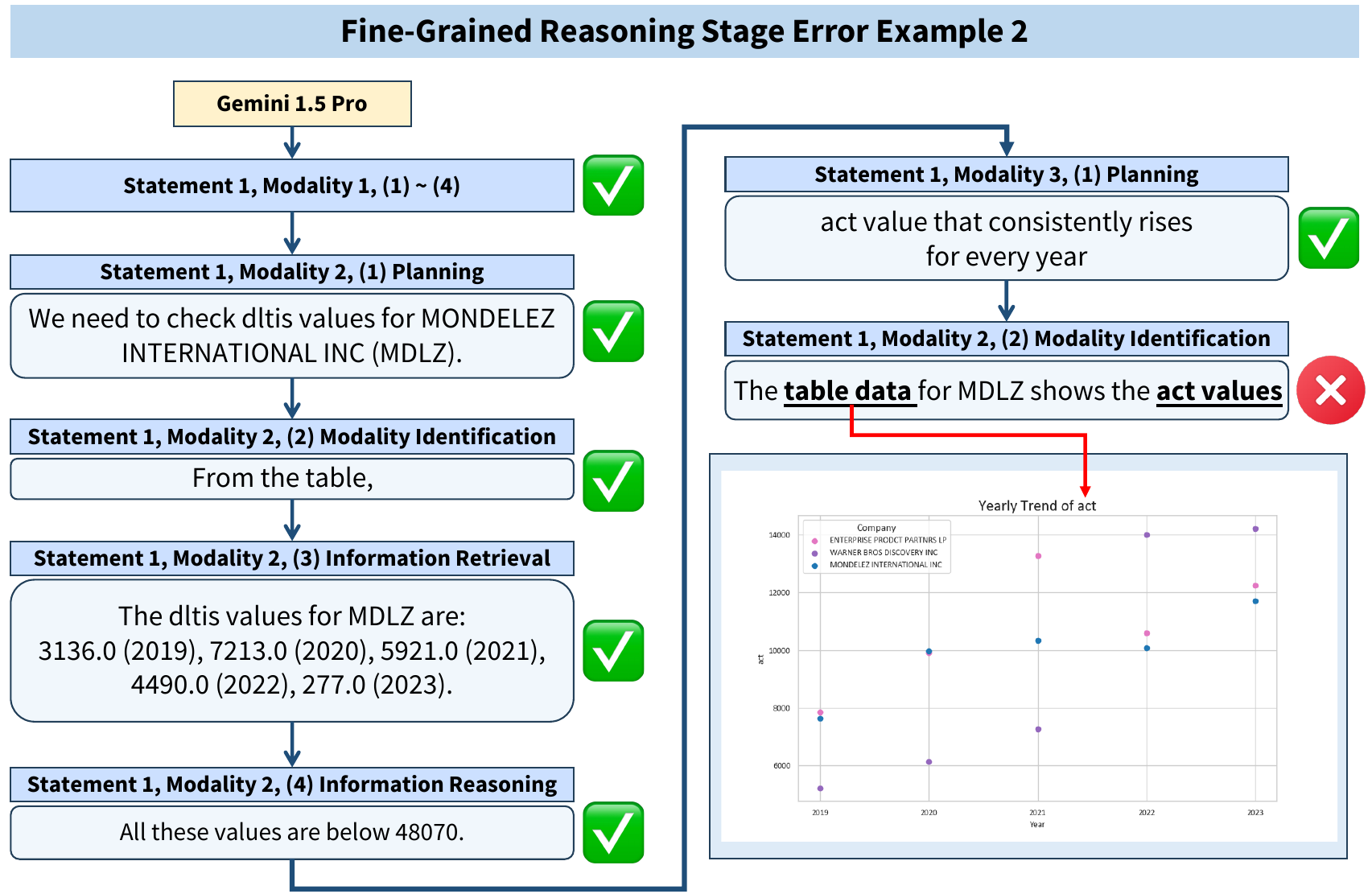} 
    \caption{An example where the Gemini 1.5 Pro succeeds in (1) Planning but fails in (2) Modality Identification. While the model successfully plans that the act value is needed, it fails by identifying the modality as Table instead of Chart, where the act value is actually presented.}
    \label{fig:appendix_reasoning_stage_error_example2} 
\end{figure*}

\begin{figure*}[htbp] 
    \centering
    \includegraphics[width=0.9\textwidth]{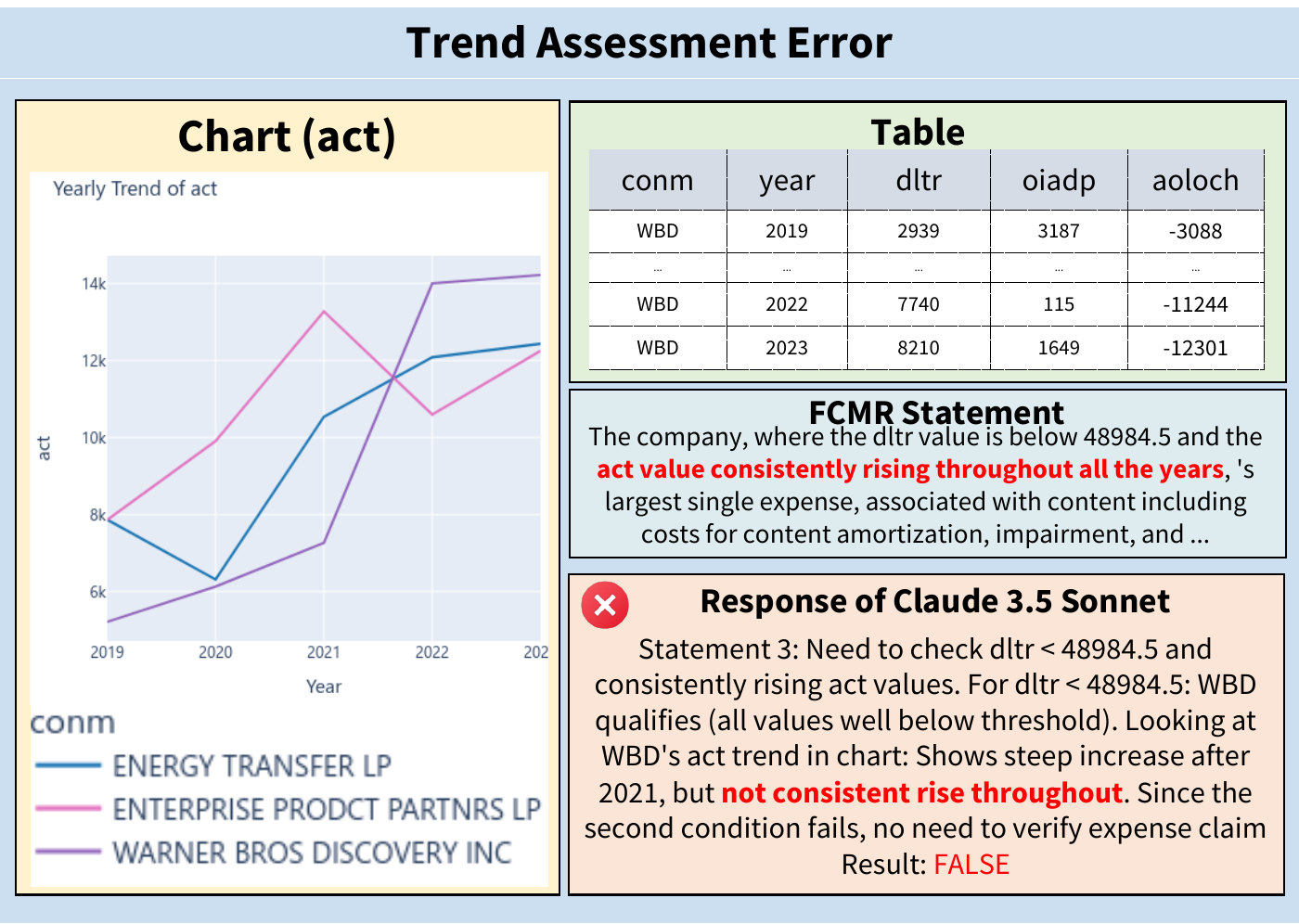} 
    \caption{In the chart, WBD shows a consistently increasing trend in act values across all years, but Claude 3.5 Sonnet fails to recognize this correctly, resulting in an error.}
    \label{fig:appendix_case_tae} 
\end{figure*}

\begin{figure*}[htbp] 
    \centering
    \includegraphics[width=0.9\textwidth]{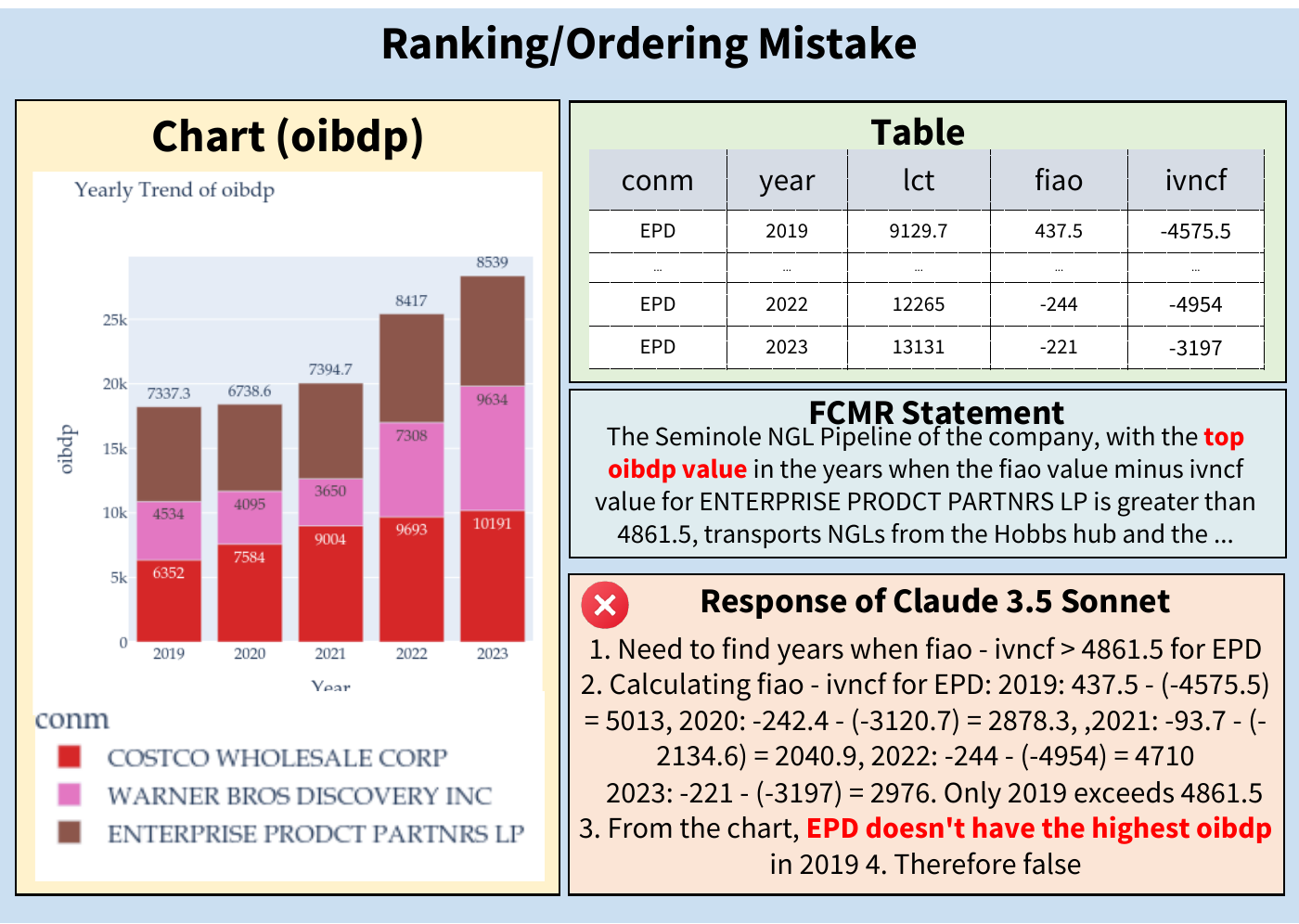} 
    \caption{An example where the Claude model fails to interpret a cumulative graph: Although the company with the highest OIBDP value in 2019 is indeed EPD (Enterprise Product Partners LP), Claude incorrectly states otherwise.}
    \label{fig:appendix_case_rom} 
\end{figure*}

\begin{figure*}[htbp] 
    \centering
    \includegraphics[width=0.9\textwidth]{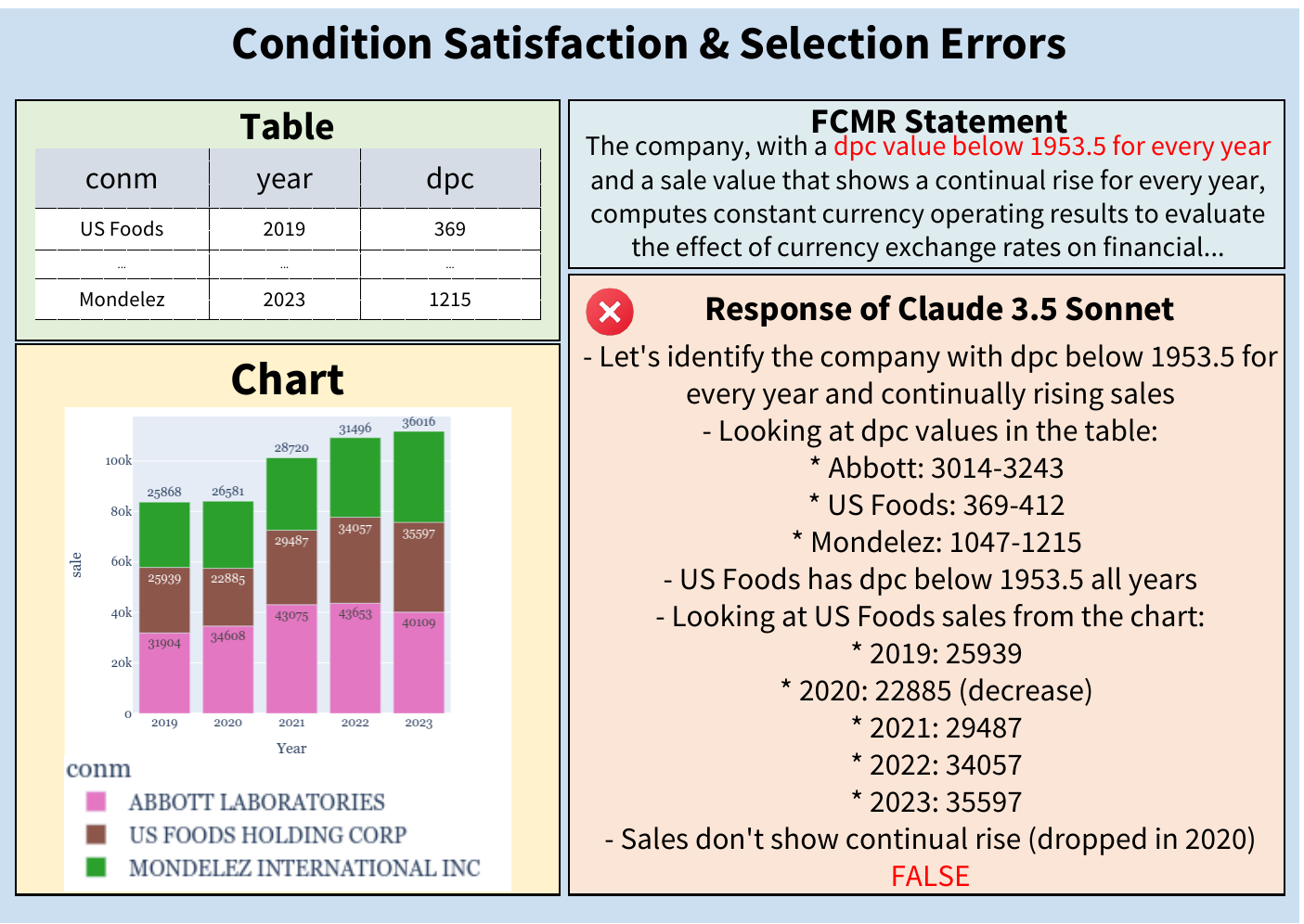} 
    \caption{The condition that the dpc value is less than 1953.5 for all years is satisfied by both US Foods and Mondelez, but Claude 3.5 Sonnet recognizes only US Foods and fails to consider Mondelez, resulting in an error.}
    \label{fig:appendix_case_csse} 
\end{figure*}

\begin{figure*}[htbp] 
    \centering
    \includegraphics[width=0.9\textwidth]{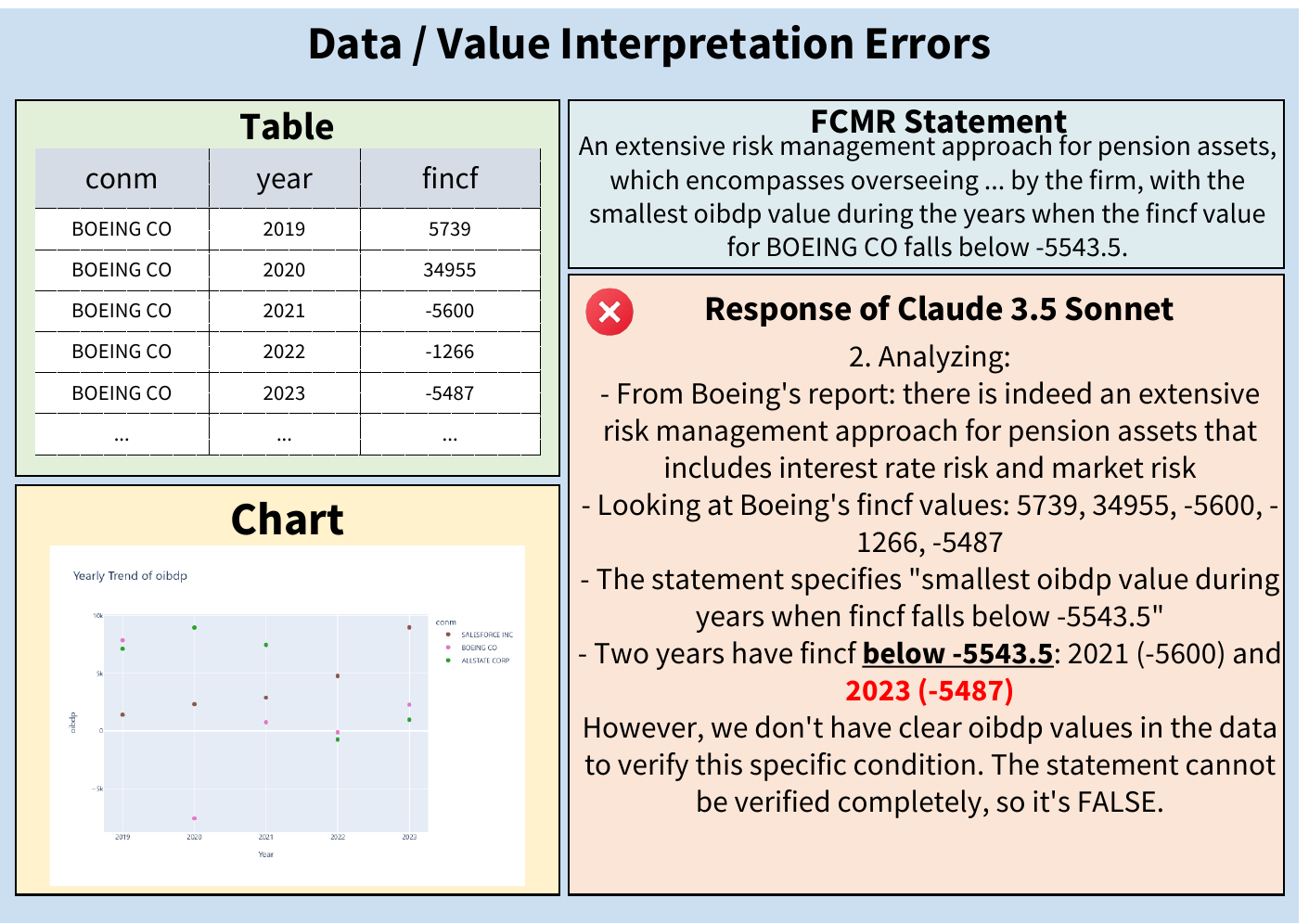} 
    \caption{-5487 is not smaller than -5543.5, but Claude failed to correctly compare the two numbers and consequently reached the wrong conclusion.}
    \label{fig:appendix_case_dvie} 
\end{figure*}

\begin{figure*}[htbp] 
    \centering
    \includegraphics[width=0.9\textwidth]{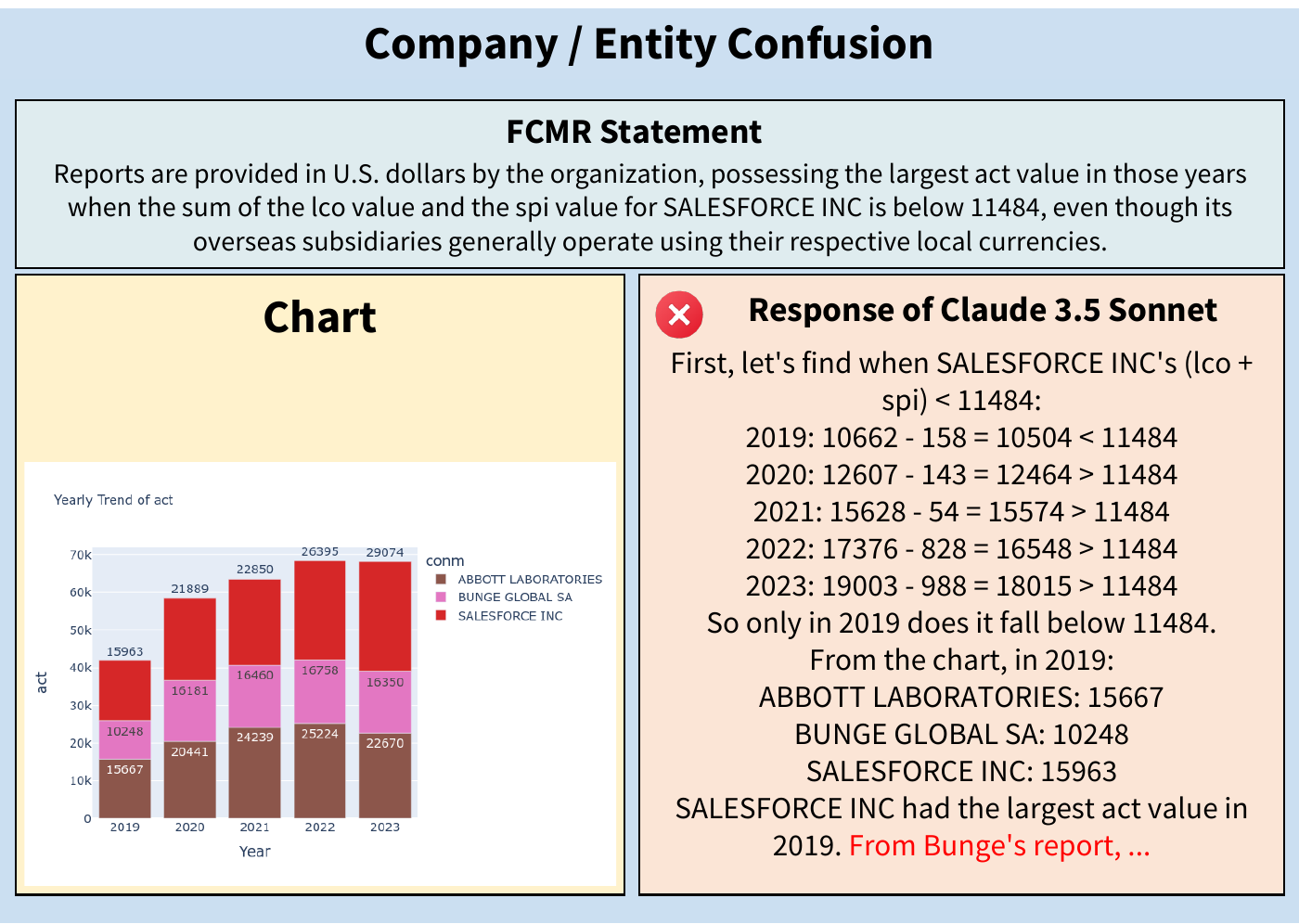} 
    \caption{Although charts and tables were successfully interpreted to determine that Salesforce had the highest act value in 2019, the company was ultimately confused in the final step by Claude, resulting in an incorrect conclusion.}
    \label{fig:appendix_case_cec} 
\end{figure*}

\begin{figure*}[htbp] 
    \centering
    \includegraphics[width=0.9\textwidth]{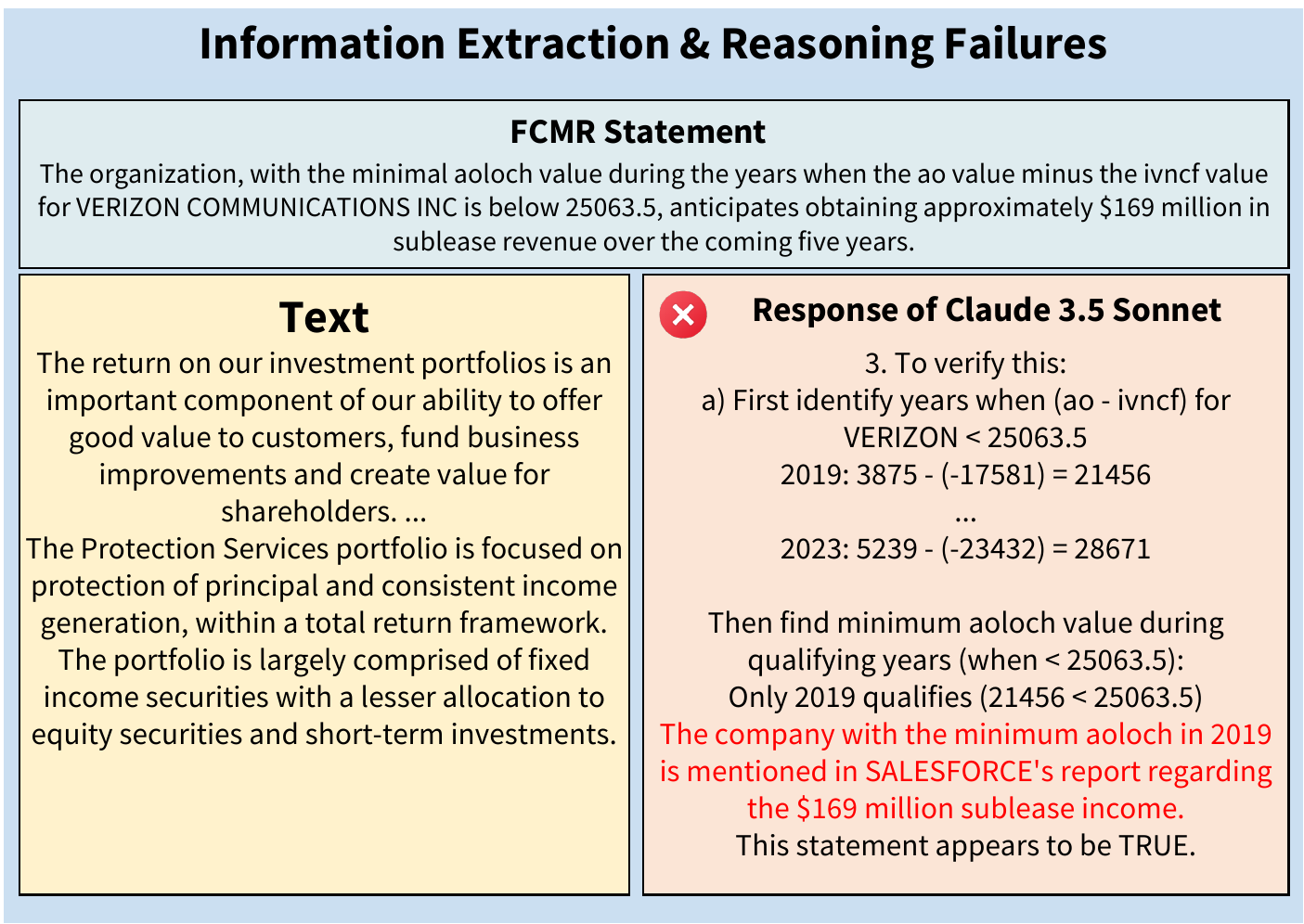} 
    \caption{To compare the aloloch value, the chart should have been considered. Instead, having relied solely on the text data, Claude led to an incorrect conclusion.}
    \label{fig:appendix_case_ierf} 
\end{figure*}

\begin{figure*}[htbp] 
    \centering
    \includegraphics[width=0.9\textwidth]{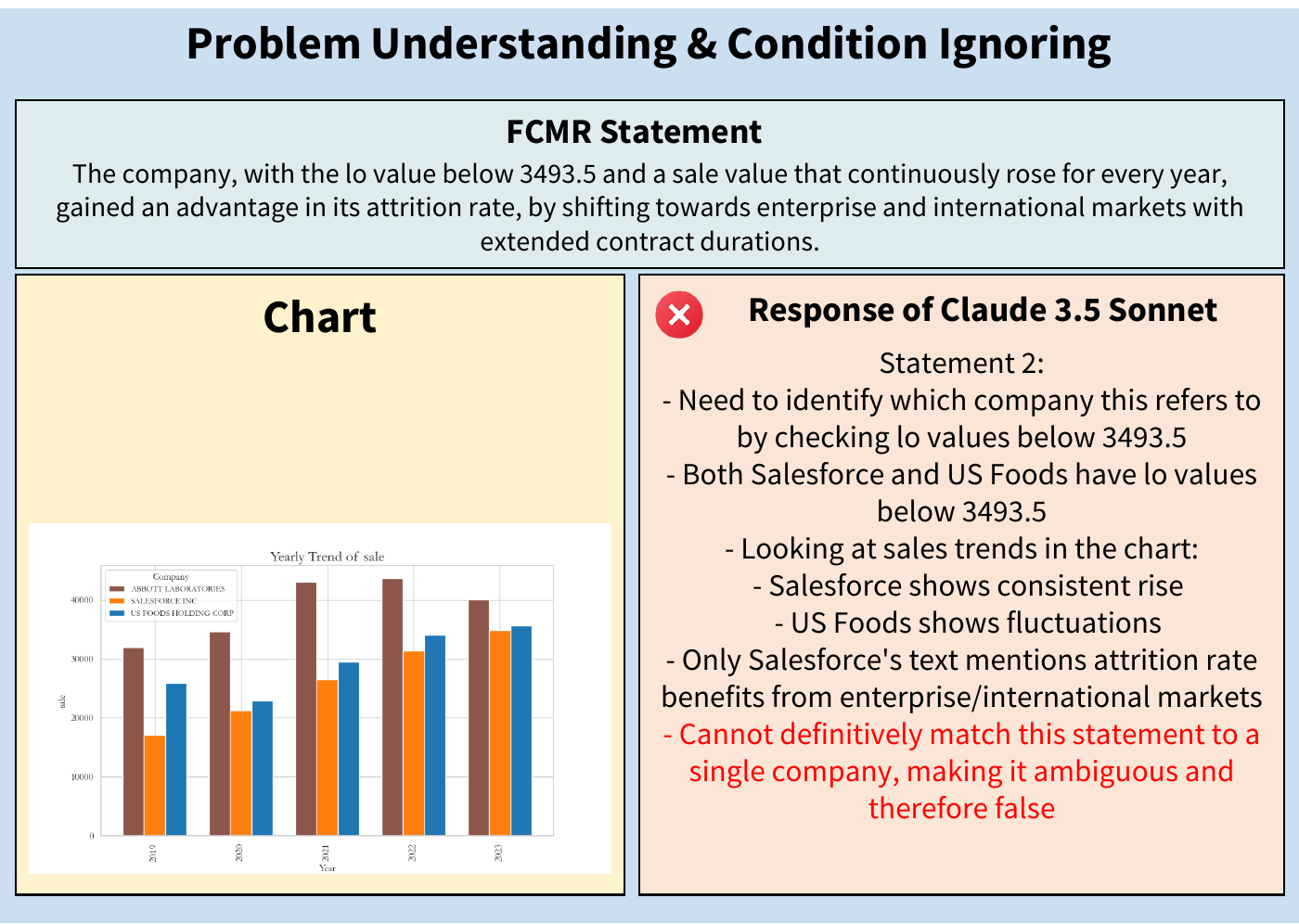} 
    \caption{Although Claude found that Salesforce satisfied all the conditions, it misunderstood the statement and consequently made an incorrect judgment.}
    \label{fig:appendix_case_puci} 
\end{figure*}

\begin{figure*}[htbp] 
    \centering
    \includegraphics[width=\textwidth]{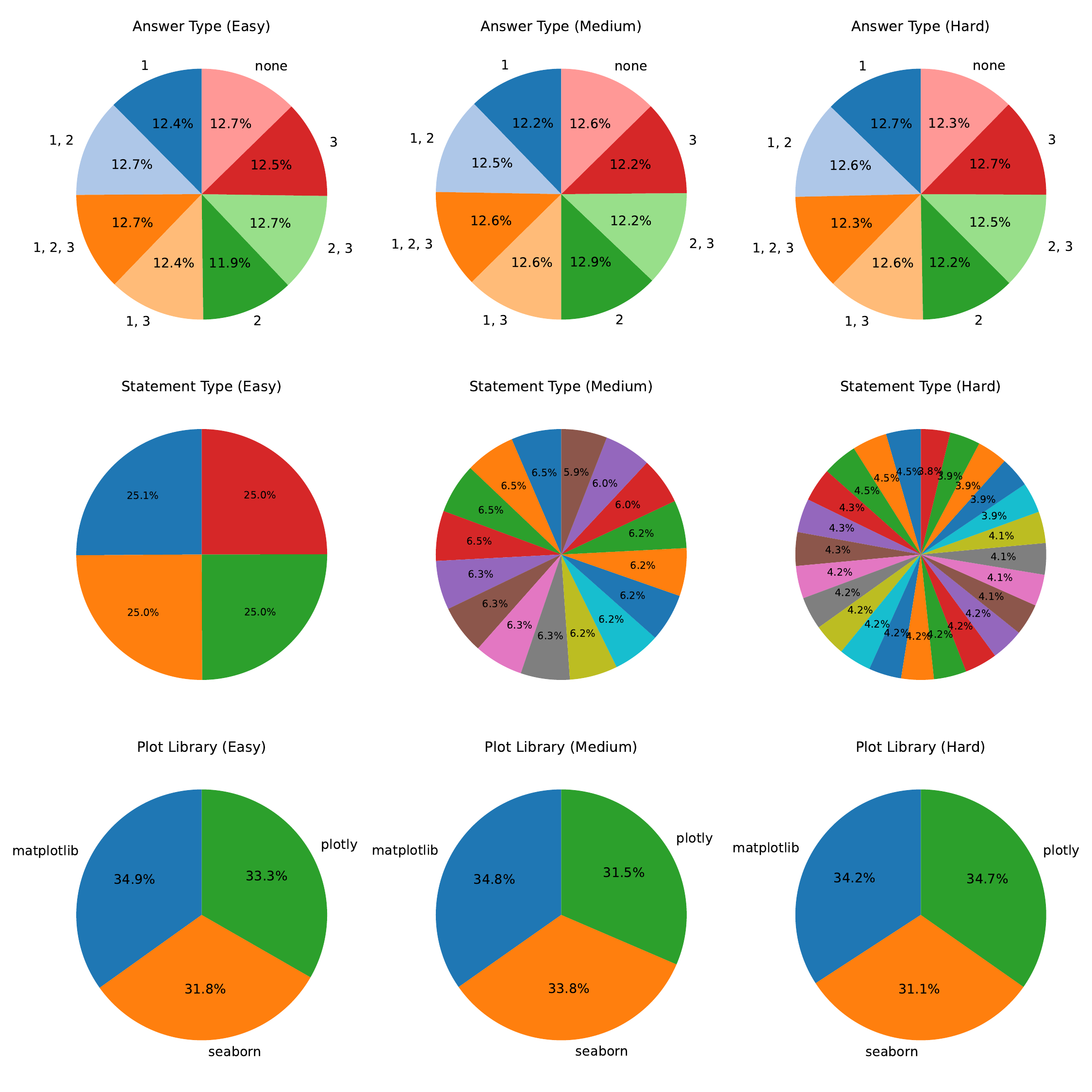} 
    \caption{Pie charts for answer types, statement types, and library usage categorized by difficulty levels.}
    \label{fig:appendix_type_pie_chart} 
\end{figure*}



\begin{figure*}[t]  
\centering
\includegraphics[width=0.98\textwidth, keepaspectratio]{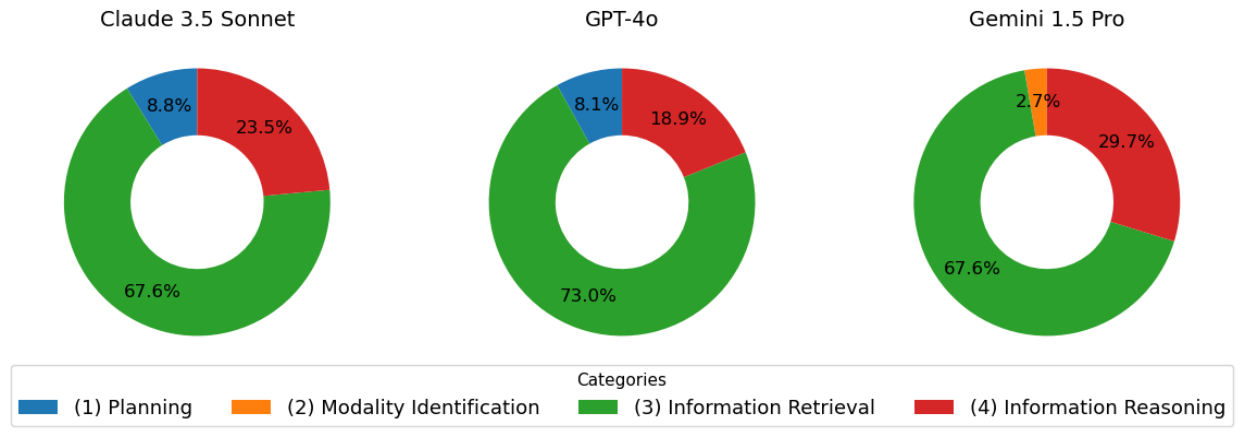}
  \caption{
  Distribution of inference failures for each MLLM across the four reasoning stages (Planning, Modality Identification, Information Retrieval, and Information Reasoning), based on manual analysis of 40 Hard-level samples.
  }
\label{fig:append_reason_infer_fail} 
\end{figure*}

\begin{table*}[t!]
\begin{center}
\normalsize 
\renewcommand{\arraystretch}{1.1}
\begin{tabular}{|l|m{13cm}|}
\hline
\textbf{Statement Types} & \textbf{Example} \\
\hline
FC & In May 2020, WORLD KINECT CORP modified and refreshed its asset-backed debt financing facility. \\ \hline
CT & The firm COSTCO WHOLESALE CORP. disclosed that its cogs values surpassed 92,765 during the year 2021. \\ \hline
AR & The company where the 2023 txt value minus the 2022 txt value equals 359.0 is WORLD KINECT CORP. \\ \hline
TR & Over the period from 2020 through 2023, SALESFORCE INC. consistently experienced an increase in its xsga values. \\ \hline
RK & During 2019, US FOODS HOLDING CORP. possessed the lowest txt value compared to other companies. \\ \hline
FC+CT & In 2023, the company whose xint values are less than 849.5 owns a terminal facility located at Fort Mifflin, which includes two docks for ships and has a total storage capacity of approximately 570 MBbls. \\ \hline
FC+AR & The company where the 2020 spi value minus the 2022 spi value equals 1036.0 recorded \$39 million and \$40 million in prior service credit amortization in 2018 and 2017, respectively. \\ \hline
FC+TR & The firm that showed a steady increase in dpc values between 2019 and 2023 must comply with the detailed regulations set by the Department of Transportation (DOT) regarding its pipeline infrastructure. \\ \hline
FC+RK & In 2022, the company, which reports fincf values greater than -2009, spreads out the amortization of its capitalized costs tied to new revenue contracts across four years. \\ \hline
CT+TR & Over the years, MONDELEZ INTERNATIONAL INC. consistently reports a fincf value exceeding -19575.5, while the ceq value demonstrates a continuous increase. \\ \hline
CT+RK & When the seq figure for WARNER BROS DISCOVERY INC falls below 10177.5, VERIZON COMMUNICATIONS INC records the lowest cogs value. \\ \hline
AR+TR & Throughout all periods, ENTERPRISE PRODCT PARTNRS LP is the company in which the cumulative nopi values surpass 1553.35, while the ao values have persistently increased. \\ \hline
AR+RK & During the years when lt value minus ibc value for UNITED PARCEL SERVICE INC is greater than 55265, the company with the lowest sale value is UNITED PARCEL SERVICE INC. \\ \hline
FC+CT+TR & Professional services are provided by the organization, which has the nopi value below 2796 for all years and the aoloch value that consistently declines for all years, to help clients with digital transformations using Salesforce solutions. \\ \hline
FC+CT+RK & The firm, with the highest act value during the years when ENERGY TRANSFER LP's intan value dips below 8059, acquired a controlling interest in USAC through a \$250 million cash transaction. \\ \hline
FC+AR+TR & The organization, with the cumulative sum of nopi values below 2840 and continuously increasing act values for every year, has provided put rights to certain consolidated subsidiaries. These put rights are omitted from the contractual obligations table due to unpredictability in payment. \\ \hline
FC+AR+RK & The business, with the minimal aoloch value in the years when the ivao value minus the ao value for VERIZON COMMUNICATIONS INC exceeds 9312.5, provides expert services to support customers in executing digital transformations leveraging Salesforce solutions. \\ \hline
\end{tabular}
\end{center}
\caption{Examples for each statement type. FC refers to Fact-Checking, CT refers to Conditional Threshold, AR refers to Arithmetic, TR refers to Trend, and RK refers to Ranking.}
\label{tab:taboptiontypeexample}
\end{table*}



\begin{table*}[t!]
\begin{center}
\normalsize 
\begin{tabular}{lp{13cm}}
\toprule
\textbf{Statement Types} & \textbf{Template} \\
\midrule
CT & The company with (column) values (greater than, less than) (threshold) in (Year) is {company}. \\ 
AR & The company where the (Year1) (column) value (plus, minus) the (Year2) (column) value equals (results) is (company). \\ 
TR & The company that showed a continuously (increasing, decreasing) trend in (column) values from (Year1) to (Year2) is (company). \\ 
RK & The company with the (highest, lowest) (column) value in (Year) is (company). \\ 
CT+TR & For all years, the company with the (column1) value (greater than, less than) (threshold) and the (column2) value continuously (increased, decreased) is (company). \\ 
CT+RK & During the years when the (column1) value for (company) is (greater than, less than) (threshold), the company with the (highest, lowest) (column2) value is (company). \\ 
AR+TR & For all years, the company with the cumulative sum of (column1) values (greater than, less than) (threshold) and the (column2) values continuously (increased, decreased) is (company). \\ 
AR+RK & During the years when (column1) value (plus, minus) (column2) value for (company1) is (greater than, less than) (threshold), the company with the (highest, lowest) (column3) value is (company2). \\ 
\bottomrule
\end{tabular}
\end{center}
\caption{Base templates of statement types. FC refers to Fact-Checking, CT refers to Conditional Threshold, AR refers to Arithmetic, TR refers to Trend, and RK refers to Ranking. In the case of Statement Types that include FC, a new template is generated by combining them with other Statement Types and Facts, where they share a common company entity.}
\label{tab:tabtemplate}
\end{table*}

\begin{table*}[t!]
\begin{center}
\centering
\normalsize
\begin{tabular}{@{}ccc@{}}
\toprule
\textbf{Difficulty} & \textbf{Modality Types} & \textbf{Statement Types} \\ \midrule
\multirow{5}{*}{Easy}   & Text               & Fact-Checking \\
                        & \multirow{2}{*}{Table} & Conditional Threshold \\
                        &                      & Arithmetic \\
                        & \multirow{2}{*}{Chart} & Trend \\
                        &                      & Ranking \\ \midrule
\multirow{8}{*}{Medium} & \multirow{2}{*}{Text + Table} & Fact-Checking + Conditional Threshold \\
                        &                      & Fact-Checking + Arithmetic \\
                        & \multirow{2}{*}{Text + Chart} & Fact-Checking + Trend \\
                        &                      & Fact-Checking + Ranking \\
                        & \multirow{4}{*}{Table + Chart} & Conditional Threshold + Trend \\
                        &                      & Conditional Threshold + Ranking \\
                        &                      & Arithmetic + Trend \\
                        &                      & Arithmetic + Ranking \\ \midrule
\multirow{4}{*}{Hard}   & \multirow{4}{*}{Text + Table + Chart} & Fact-Checking + Conditional Threshold + Trend \\
                        &                      & Fact-Checking + Conditional Threshold + Ranking \\
                        &                      & Fact-Checking + Arithmetic + Trend \\
                        &                      & Fact-Checking + Arithmetic + Ranking \\ \bottomrule
\end{tabular}
\end{center}
\caption{Detailed Statement Types by Difficulty and Modality Types. For the Easy level, all three answer statements are single-modal one-hop, while for the Medium level, all three statements are cross-modal two-hop. At the Hard level, all three statements consist of cross-modal three-hop.
For Easy level questions, while each of the three statements is single-modal one-hop, determining the correct overall answer for the instance (i.e., selecting all true statements) requires processing information from all three provided modalities (text, table, and chart), ensuring comprehensive multi-modal engagement.
Specific examples can be found in Table \ref{tab:taboptiontypeexample}.
}
\label{tab:option_types_diff}
\end{table*}

\begin{figure*}[t]  
\centering
\includegraphics[width=0.95\textwidth, keepaspectratio]{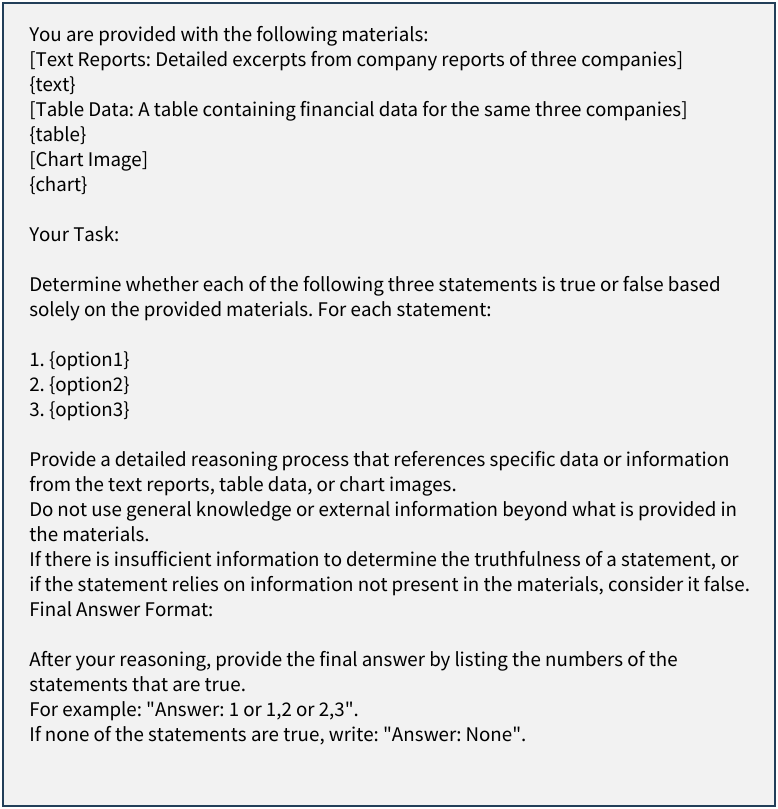}
  \caption{MLLMs zero-shot prompt.}
  \label{fig:figappendixprompt0shot}
\end{figure*}

\begin{figure*}[t]  
\centering
\includegraphics[width=0.95\textwidth, keepaspectratio]{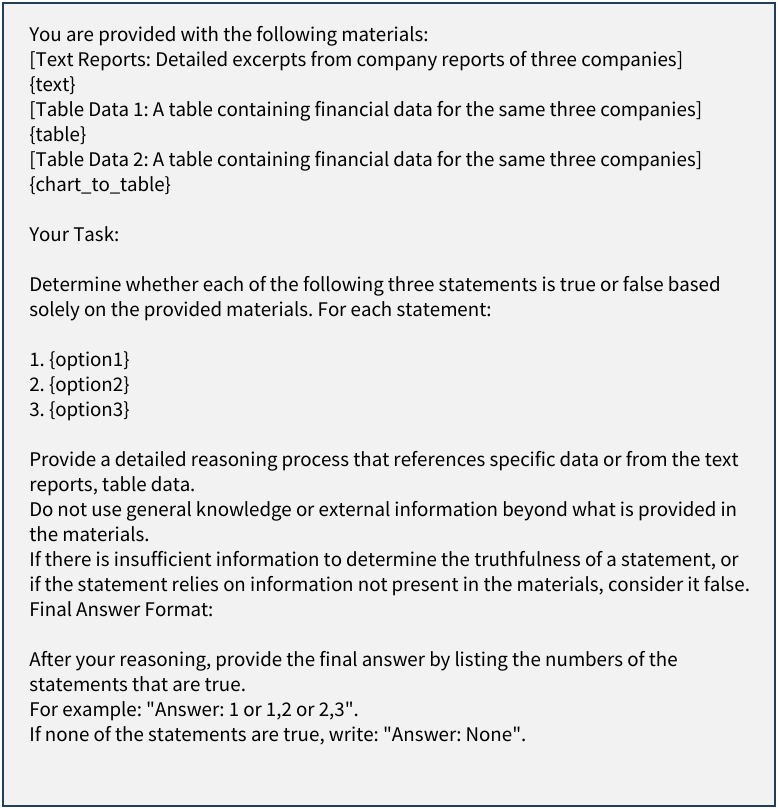}
  \caption{(M)LLMs + Deplot zero-shot prompt.}
  \label{fig:deplot_figappendixprompt0shot}
\end{figure*}

\begin{figure*}[t]  
\centering
\includegraphics[width=0.98\textwidth, keepaspectratio]{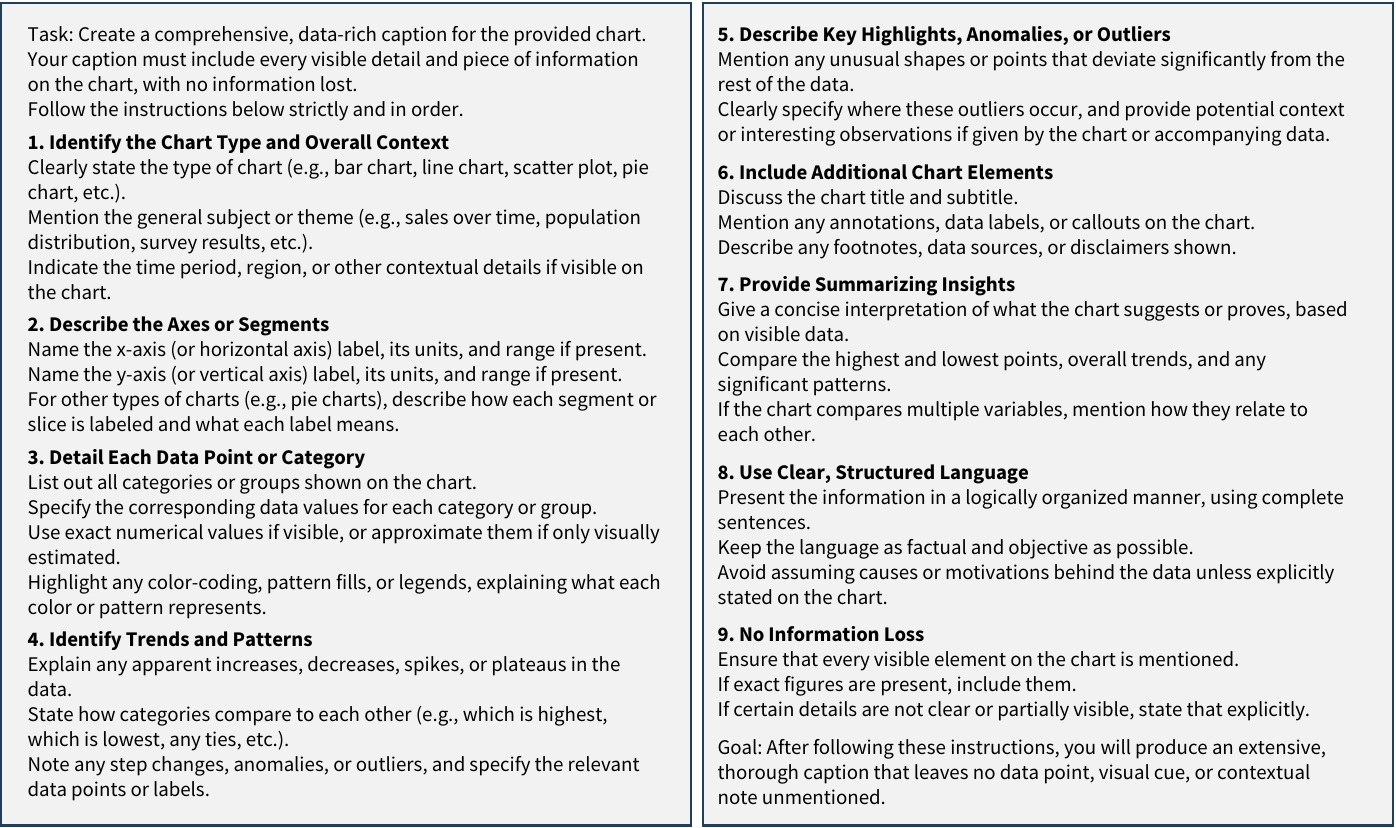}
  \caption{Modality Integration prompt.}
  \label{fig:modality_integ_prompt_figure}
\end{figure*}

\begin{figure*}[t]  
\centering
\includegraphics[width=0.95\textwidth, keepaspectratio]{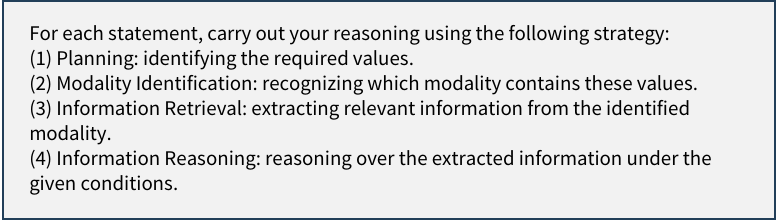}
  \caption{4-Stage Reasoning Strategy prompt.}
  \label{fig:four_stage_reasoning_figure}
\end{figure*}

\begin{figure*}[t]  
\centering
\includegraphics[width=0.95\textwidth, keepaspectratio]{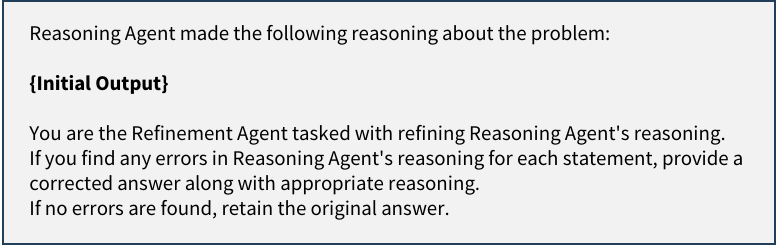}
  \caption{Self-Refine prompt.}
  \label{fig:self_refine_prompt_figure}
\end{figure*}

\begin{figure*}[t]  
\centering
\includegraphics[width=0.95\textwidth, keepaspectratio]{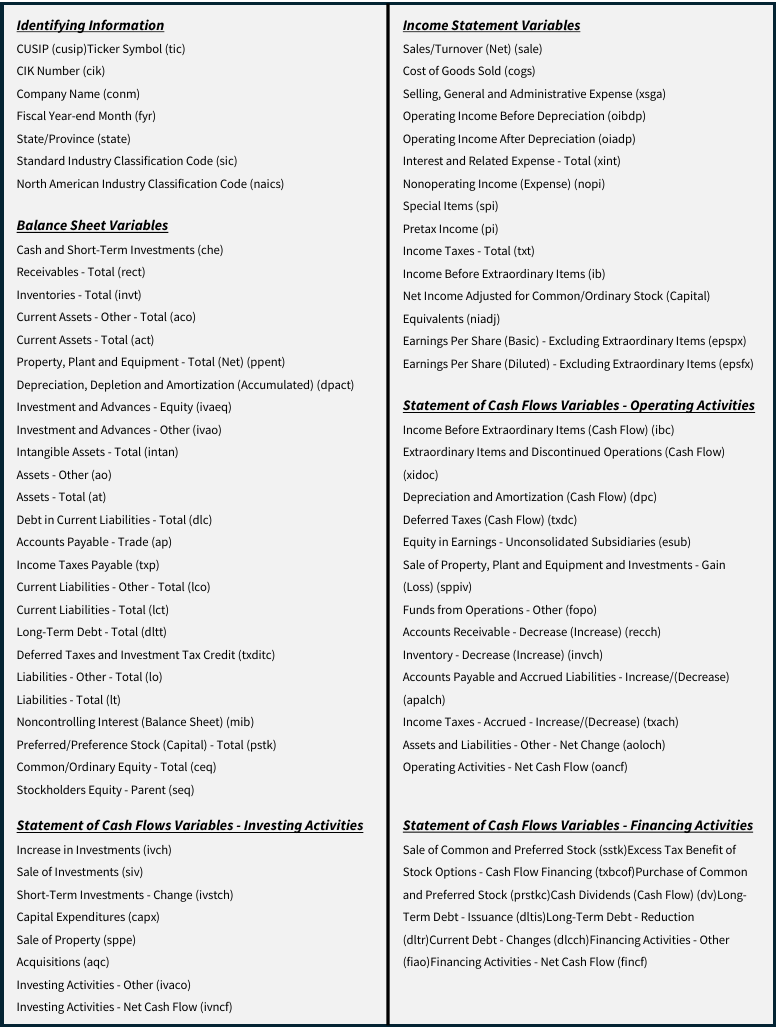}
  \caption{Description of each column in the Annual Simplified Financial Statement.}
  \label{fig:figcolumn}
\end{figure*}
\end{document}